\documentclass[
]{ceurart}

\usepackage{listings}
\usepackage{multirow}
\usepackage{tablefootnote}
\usepackage{graphicx, subcaption, ragged2e}
\usepackage{amsmath}
\usepackage{multicol}
\usepackage{makecell}
\usepackage{threeparttable}
\usepackage{pifont}
\usepackage{ulem}

\newcommand{\cmark}{\ding{51}}
\newcommand{\xmark}{\ding{55}}

\begin{document}

\copyrightyear{2025}
\copyrightclause{Copyright for this paper by its authors.
  Use permitted under Creative Commons License Attribution 4.0
  International (CC BY 4.0).}

\conference{GENNEXT@SIGIR'25: The 1st Workshop on Next Generation of IR and Recommender Systems with Language Agents, Generative Models, and Conversational AI, Jul 17, 2025, Padova, Italy}

\title{FACap: A Large-scale Fashion Dataset for Fine-grained Composed Image Retrieval}


\author[1,2]{François Gardères}[%
email=francois.garderes@louisvuitton.com,
]
\address[1]{Louis Vuitton}
\address[2]{Inria, \'Ecole normale sup\'erieure, CNRS, PSL Research University}

\author[2]{Shizhe Chen}[%
]

\author[1]{Camille-Sovanneary Gauthier}[%
]

\author[2,3]{Jean Ponce}[]
\address[3]{Courant Institute of Mathematical Sciences and Center for Data Science, New York University}



\begin{abstract}
The composed image retrieval (CIR) task is to retrieve target images given a reference image and a modification text. 
Recent methods for CIR leverage large pretrained vision-language models (VLMs) and achieve good performance on general-domain concepts like color and texture.
However, they still struggle with application domains like fashion, because the rich and diverse vocabulary used in fashion requires specific fine-grained vision and language understanding. An additional difficulty is the lack of large-scale fashion datasets with detailed and relevant annotations, due to the expensive cost of manual annotation by specialists.
To address these challenges, we introduce  \textbf{FACap}, a large-scale, automatically constructed fashion-domain CIR dataset. It leverages web-sourced fashion images and a two-stage annotation pipeline powered by a VLM and a large language model (LLM) to generate accurate and detailed modification texts.
Then, we propose a new CIR model \textbf{FashionBLIP-2}, which fine-tunes the general-domain BLIP-2 model on FACap with lightweight adapters and multi-head query-candidate matching to better account for fine-grained fashion-specific information. 
FashionBLIP-2 is evaluated with and without additional fine-tuning on the Fashion IQ benchmark and the enhanced evaluation dataset enhFashionIQ, leveraging our pipeline to obtain higher-quality annotations.
Experimental results show that the combination of FashionBLIP-2 and pretraining with FACap significantly improves the model's performance in fashion CIR especially for retrieval with fine-grained modification texts, demonstrating the value of our dataset and approach in a highly demanding environment such as e-commerce websites.
Code is available at \url{https://fgxaos.github.io/facap-paper-website/}.
\end{abstract}

\begin{keywords}
  Composed Image Retrieval \sep
  Multimodal Fusion \sep
  Fashion Domain
\end{keywords}

\maketitle

\section{Introduction}

\begin{figure}[tbh]
\centering
    \begin{subfigure}[b]{0.8\linewidth}
    \centering
    \includegraphics[width=\linewidth]{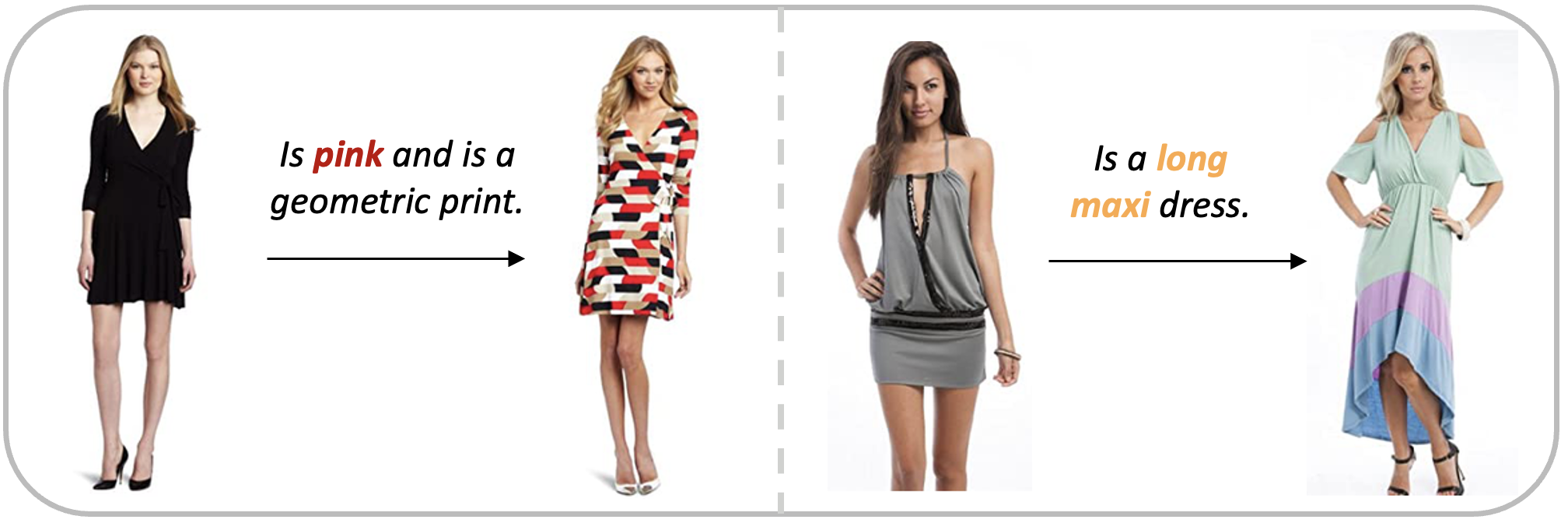}
    \caption{Examples from the FashionIQ dataset~\cite{wu2020fashioniq}. {\normalfont Left: incorrect annotation (the target dress is not pink). Right: vague annotation lacking sufficient details to accurately retrieve the target image, like color or shape.}}
    \label{fig:teaser-fashioniq}
    \end{subfigure}
\hfill
    \begin{subfigure}[b]{0.8\linewidth}
    \centering
    \includegraphics[width=\linewidth]{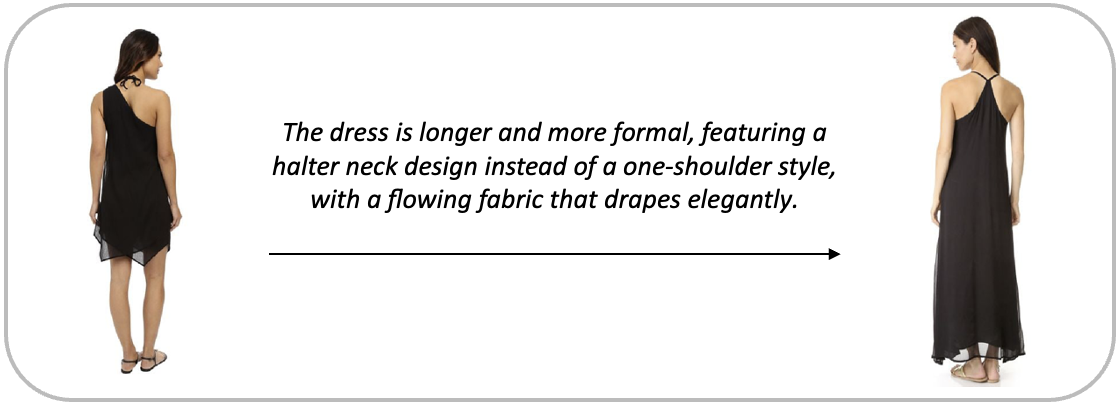}
    \caption{Example from our FACap dataset.}
    \label{fig:teaser-facap}
    \end{subfigure}

    \caption{Our automatically constructed FACap dataset offers more detailed and accurate annotations than existing datasets for the fashion CIR task.}
    \label{fig:teaser}
\end{figure}

Efficiently retrieving fashion images based on user preferences is crucial for enhancing e-commerce experience, from online shopping to inspiration and brand discovery.
The preferences relate both to search interaction preferences \textemdash  querying with images for instance \textemdash  and taste and vocabulary preference to really adapt to user needs. 
Traditional image-to-image~\cite{shao2023global} or text-to-image~\cite{rao2022does} retrieval methods primarily support single-modality queries and fall short in handling more complex, real-world scenarios. For instance, a user may want to find a product similar to the one they have seen before, but with specific changes, like a different color, style, or feature.
To address this, recent works have increasingly focused on composed image retrieval (CIR)~\cite{wu2020fashioniq,vo2019composing}, which aims to retrieve relevant fashion images by leveraging a reference image along with a modification text that describes specific alterations. 

Most existing methods for Fashion CIR~\cite{baldrati2022conditioned,liu2024bi,feng2024improving,liu2024candidate,zhao2022progressive,zhao2024unifashion} fine-tune pretrained vision-language models (VLMs) like CLIP~\cite{radford2021clip} or BLIP-2~\cite{li2023blip2} to map images and texts into a shared multimodal space. The embeddings of the reference image and modification text are then fused and compared with the embeddings of candidate images to identify the most relevant match.
However, these approaches are constrained by the limitations of current Fashion CIR datasets.
For example, Fashion IQ~\cite{wu2020fashioniq}, a widely used dataset for this task, is limited in scale, containing only 18k \textit{<reference image, modification text, target image>} triplets across just three fashion categories: dresses, shirts, and tops. A larger scale dataset would be able to better represent the concept diversity of fashion-domain knowledge.
Furthermore, the crowdsourced captions in Fashion IQ are short, noisy and lack details, as shown in Figure~\ref{fig:teaser-fashioniq} and confirmed by our quality evaluation in Table~\ref{tab:dataset_quality_eval}. 
A better CIR experience is expected from three features of the modification text: faithfulness, levels of detail and discriminative power. Reaching a high level of quality is time-consuming and expensive, as it requires to manually annotate a large number of CIR triplets. 
The noisy and limited data available today hinder the existing models' ability to understand fine-grained fashion-related features crucial for fashion search tasks, such as specific collar types or textures.

To tackle the data scarcity challenge, some approaches have attempted to increase the data size, for example by generating reverse descriptions~\cite{liu2024bi} for reference and target images; but generating difference descriptions for two images~\cite{park2019robust} is itself a challenging task.
Other approaches attempt to eliminate the need for training data by performing zero-shot CIR~\cite{baldrati2023zero,saito2023pic2word,tang2024context,yang2024ldre,lin2024fine,karthik2023vision} with the help of pretrained VLMs, but their performance suffers from the absence of domain-specific representation learning.
More recent efforts pretrain VLMs using web-crawled fashion images to learn more accurate multimodal representations for fashion~\cite{zhao2022progressive,zhao2024unifashion}, but raw web data are often noisy and lack the necessary comparisons between pairs of images for effective CIR training. 
 
In this work, we introduce Fashion Automatic Caption (FACap), a large-scale fashion-domain CIR dataset with fine-grained annotations.
FACap automatically pairs web-sourced fashion images and employs a two-stage annotation pipeline to generate modification texts, hence creating CIR triplets.
The first stage refines original noisy web captions using a VLM to produce long, faithful, and detailed descriptions for each image. 
The second stage utilizes a large language model (LLM) to analyze the differences between reference and target image captions, generating concise and accurate modification texts.
With over 227k CIR triplets, FACap offers a high-quality dataset addressing the challenges of scale, accuracy, and detail in fashion CIR, as evidenced in our quality evaluation.
We also introduce the FashionBLIP-2 model for Fashion CIR task using BLIP-2~\cite{li2023blip2} as backbone, and lightweight adapter modules to specialize it for fashion retrieval needs.
Additionally, instead of relying on global features to match the multimodal query and candidate image, we design a multi-head query-candidate matching method that uses multiple feature representations to capture more fine-grained details.
We evaluate the performance of our model in two settings: with and without fine-tuning on downstream fashion datasets.

Experimental results demonstrate that pretraining on FACap significantly improves model performance for Fashion CIR and showcase the effectiveness of our FashionBLIP-2 model.

To summarize, our contribution is three-fold:
\begin{itemize}
    \item We propose an automatic data construction method to scale up Fashion CIR datasets with web-sourced images and foundation models, resulting in a large-scale and high-quality dataset FACap.

    \item We propose the FashionBLIP-2 model, incorporating BLIP-2 with lightweight adapters for fashion domain adaptation and multi-head matching to cover fine-grained details. 

    \item Experimental results on two benchmarks with and without downstream fine-tuning demonstrate the value of FACap and our model. 

\end{itemize}

\section{Related Work}

\begin{figure*}[t]
    \centering
    \includegraphics[width=\linewidth]{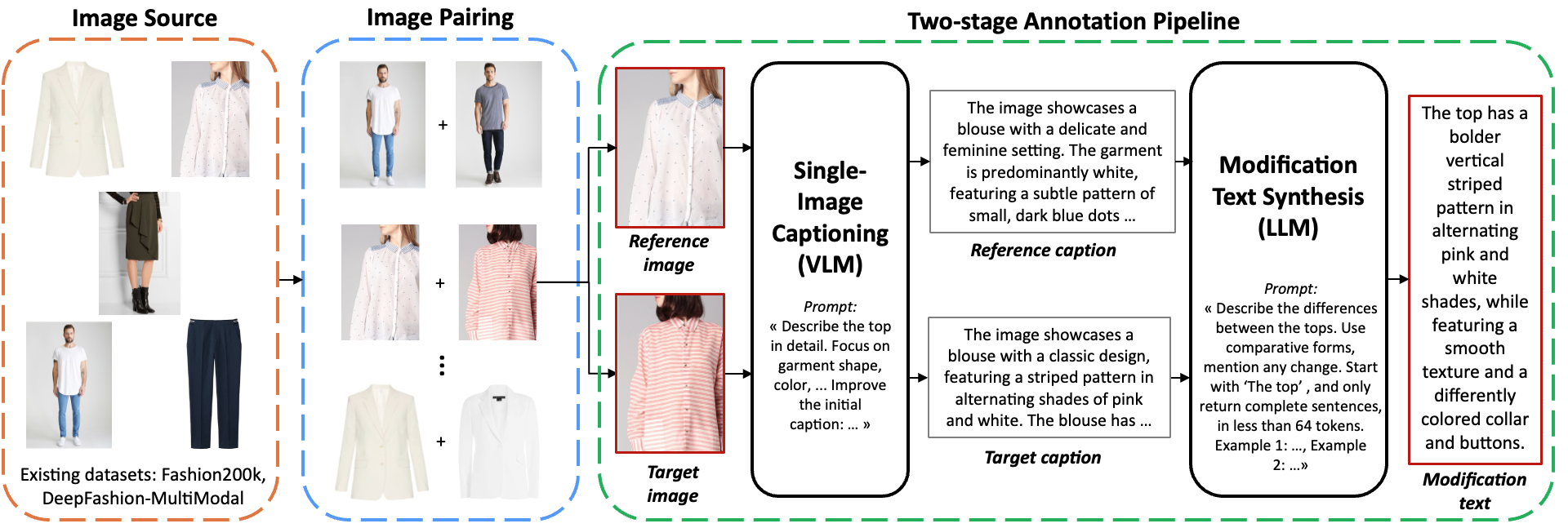}
    \caption{The proposed data construction pipeline to automatically generate CIR triplets. The images are extracted from large existing fashion datasets, then paired based on their visual similarity with images from the same product category. Then, our two-stage annotation process captions the images with a VLM, and an LLM generates a synthetic description of the changes applied on the reference image to obtain the target image.}
    \label{fig:datagen_pipeline}
\end{figure*}

\subsection{Composed Image Retrieval}

Existing approaches~\cite{baldrati2022conditioned,liu2024bi,feng2024improving,liu2024candidate,zhao2022progressive,zhao2024unifashion} to composed image retrieval (CIR) mainly focus on learning a joint representation of the reference image and the modification text.
The CLIP4CIR~\cite{baldrati2022conditioned} model leverages CLIP~\cite{radford2021clip} to encode images and texts and then uses MLPs to aggregate embeddings of the two modalities.
To further enhance the modality representation, recent works~\cite{bai2024sentence,liu2024bi} have employed more powerful pretrained multimodal models such as BLIP~\cite{li2022blip} and BLIP-2 \cite{li2023blip2}, yielding significant performance improvements.
However, these methods rely on a single global vector for representation, which limits their ability to capture fine-grained details.
To improve fine-grained CIR, TG-CIR~\cite{wen2023target} introduces both global and local attribute features with orthogonal regularization to learn more independent attribute features.
ARTEMIS~\cite{delmas2022artemis} and CaLa~\cite{jiang2024cala} propose two auxiliary methods leveraging
image-text interactions in the CIR triplet to enhance query-target matching.
Liu et al.~\cite{liu2024candidate} employ a two-stage approach, where the first stage uses a single global feature to filter out easy negatives, and the second stage leverages a dual-encoder architecture for fine-grained re-ranking.
SPRC~\cite{bai2024sentence} proposes an additional sentence-level prompt for image-text fusion and a text prompt alignment loss to improve prompt learning.

One key challenge in CIR is the lack of high-quality supervised data. Existing CIR datasets, such as Fashion IQ \cite{wu2020fashioniq}, CIRR \cite{liu2021cirr} and CIRCO \cite{baldrati2023zero}, are significantly smaller than broader vision-language datasets like COCO \cite{lin2014microsoft} and LAION-5B \cite{schuhmann2022laion5b}. 
To address this limitation, a line of work focuses on zero-shot CIR (ZS-CIR) \cite{baldrati2023zero,saito2023pic2word,tang2024context,yang2024ldre,lin2024fine,karthik2023vision}, aiming to develop generalized CIR models without the need for annotated data. ZS-CIR methods typically translate an image into text with a captioning model or textual inversion~\cite{galimage}. Yet, the performance gap between zero-shot methods and fully domain-adapted ones remains significant.
Another type of approaches explores data augmentation~\cite{liu2024bi} and synthetic data generation~\cite{gu2023compodiff,feng2024improving,levy2024data,ventura2024covr}.
BLIP4CIR+Bi \cite{liu2024bi} extends CIR datasets by adding reverse triplets, but results in less specific and less accurate modification texts.
CompoDiff~\cite{gu2023compodiff} uses an LLM to create new modification texts and generates the corresponding target images with a diffusion model\cite{rombach2022stablediffusion,hertz2022prompttoprompt}. However, limitations in image generation models compromise the faithfulness and quality of the generated images.
SPN~\cite{feng2024improving}, LaSCo~\cite{levy2024data}, and CoVR-2 \cite{ventura2024covr} instead only leverage VLMs and LLMs to generate modification texts for paired real images or videos.
But these datasets focus more on general-domain images and fail to capture the fine-grained, fashion-specific vocabulary and visual details critical for fashion CIR tasks.

To improve fashion-domain CIR, recent efforts have focused on improving the pretraining of large multimodal models on fashion images.
FashionViL~\cite{han2022fashionvil} proposes a multi-view contrastive learning approach and pseudo-attributes classification to improve representation learning with fashion image-text pairs.
Zhao et al.~\cite{zhao2022progressive} proposes a progressive learning strategy, transitioning from general-domain pretraining to fashion domain pretraining.
FAME-ViL~\cite{han2023fame} uses multi-task learning on heterogeneous fashion tasks, while UniFashion~\cite{zhao2024unifashion} further extends pretraining fashion datasets and tasks to include a broader range of multimodal generation and retrieval tasks, achieving state-of-the-art results in fashion CIR benchmarks. 
Nevertheless, existing fashion-focused pretraining mainly relies on image-text pairs rather than CIR triplets due to the scarcity of annotated triplets, limiting the model's ability to learn comparisons between two images.
In this work, we address this gap by designing an automatic method to generate CIR triplets from fashion-domain images, and enhance pretraining efficiency for fashion CIR.

\subsection{Large vision and language models}

Recently, large language models (LLMs) like GPT~\cite{brown2020language} and LLaMA~\cite{dubey2024llama} have achieved remarkable success on various textual tasks like text generation and reasoning.
Building on this foundation, numerous models have been developed to extend LLMs with visual perception capabilities by encoding images as inputs to the LLMs, resulting in powerful large vision-language models (VLMs) like BLIP-2~\cite{li2023blip2}, LLaVA~\cite{liu2024visual}, GPT-4V~\cite{achiam2023gpt}, InternVL ~\cite{chen2024internvl} and many more~\cite{wang2022git,gan2022vision}. 
These VLMs effectively combine textual and visual information and have set new benchmarks across diverse tasks such as image captioning~\cite{lin2014microsoft}, visual question answering~\cite{antol2015vqa} and so on.

\begin{figure}
    \centering
    \begin{subfigure}[t]{0.45\textwidth}
        \includegraphics[width=\linewidth]{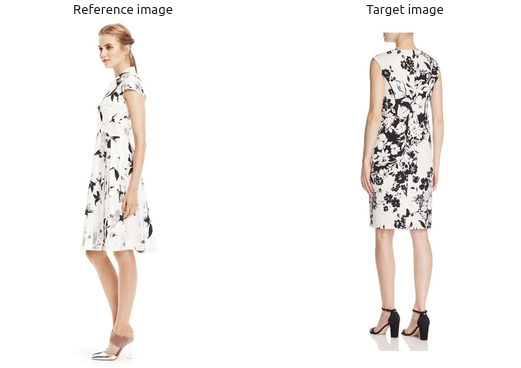}
        \caption{The dress is sleeker and more form-fitting, with a round neckline and a sleeveless design, featuring a denser floral pattern in black and white on a white base.}
    \end{subfigure}
    \hfill
    \begin{subfigure}[t]{0.45\textwidth}
        \includegraphics[width=\linewidth]{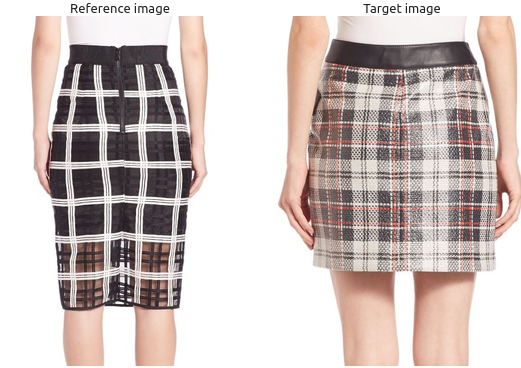}
        \caption{The skirt is shorter and more form-fitting with a mini length, featuring a classic tartan plaid pattern in black, white, and red, and a smooth woven texture instead of the layered effect.}
    \end{subfigure}
    \caption{Examples from the FACap dataset. The caption of each image pair corresponds to their modification text.} 
    \label{fig:facap_data_examples}
\end{figure}

\begin{table*}[t]
\centering
\caption{
    Comparison of different datasets.  We exclude certain web image sources to avoid licensing constraints, resulting in fewer unique images than the FACAD dataset~\cite{Xuewen2020FACAD}. FACAD also includes noisy web descriptions with unstandardized language, leading to a larger vocabulary size. Instead, the captions in our FACap are automatically cleaned and contain more details.
}
\label{tab:data_stats}
\begin{tabular}{lccccccc} \toprule
 & \#Uniq imgs & Ann. type & Pair type & \#Pairs & Vocab size & Avg. length \\ 
\midrule
MSCOCO~\cite{lin2014microsoft} & 328,000 & Manual & <img, caption> & 1,640,000 & 26,848 & 10.5 \\ 
\midrule
FACAD~\cite{Xuewen2020FACAD}  & 993,000 & Web & <img, caption> & 130,000 & 15,807 & 21 \\
Fashion IQ~\cite{wu2020fashioniq} & 25,136 & Manual & \makecell{<ref img, mod txt, \\ tgt img>} & 18,000 & 4,401 & 6.36 \\ \midrule
\multirow{2}{*}{\textbf{FACap (Ours)}} & \multirow{2}{*}{227,680} & \multirow{2}{*}{Auto} & \makecell{<ref img, mod txt, \\ tgt img>} & 227,680 & 9,273 & 23.38 \\
 &  &  & <img, caption> & 227,680 & 18,689 & 82.90 \\ \bottomrule
\end{tabular}
\end{table*}

While most VLMs are designed to process single-image inputs, recent advancements~\cite{liu2024llavanext,li2024llavaonevision,li2024llavanextinterleave,achiam2023gpt} have aimed to improve multi-image capabilities. 
However, this progress introduces two key challenges. 
First, on the model side, handling multiple images significantly increases the token count, leading to issues with context length. To address this, various image token compression techniques~\cite{shen2024longvu,li2024tokenpacker} have been proposed for VLMs. 
Second, on the data side, multi-view image datasets~\cite{liu2024llavanext} remain limited, restricting the ability of current VLMs to excel in multi-image reasoning tasks.
In this work, instead of directly using VLMs to generate modification texts for image pairs, we propose a two-stage pipeline that leverages the strengths of VLMs for detailed single-image captioning and LLMs for advanced text reasoning. This approach ensures high-quality annotations that precisely capture the fine-grained details essential for fashion CIR.

\section{The Fashion Automatic Caption Dataset}

To tackle the data scarcity challenge in fashion-domain CIR, we introduce a large-scale Fashion Automatic Caption dataset (FACap), generated automatically using existing fashion image datasets and foundation models. It provides detailed image captions and CIR triplets with both global and fashion-specific vocabulary, so that CIR methods can leverage precise fine-grained textual and visual concepts.

\subsection{Dataset Construction}
\label{sec:data_collection_method}

Our goal is to generate triplets of the form \textit{<reference image, modification text, target image>} for Fashion CIR.
Figure~\ref{fig:datagen_pipeline} illustrates the automatic data construction pipeline, including image source collection, image pairing, and our two-stage annotation using single-image captioning and modification text generation.

\noindent\textbf{Image sources.}
We use two publicly available fashion datasets: Fashion200k~\cite{han2017automatic} and DeepFashion-MultiModal~\cite{jiang2022text2human}, both originally crawled from online shopping websites.
The Fashion200k dataset comprises approximately 200k images across five categories: dresses, jackets, pants, skirts, and tops. The images are accompanied by product descriptions, which, while useful, tend to be noisy.
DeepFashion-MultiModal~\cite{jiang2022text2human} is a refined version of the DeepFashion~\cite{liu2016deepfashion} dataset. It consists of 44,096 high-resolution model-worn images of clothing, each annotated with automatically-parsed attributes from product descriptions and manually-labeled shape and texture information.
Note that images in both datasets are distinct from those used in the downstream datasets, ensuring there is no information leakage.

\noindent\textbf{Image pairing.}
From this large image pool, we extract pairs of images to create a list of reference and target image pairs for the CIR task. 
We constrain the visual similarity of the image pairs:
if two images are too different, the modification text will focus on describing the target image, ignoring the reference image. On the other hand, if two images are overly similar, it can be challenging for automatic systems to describe their subtle differences.
To address this similarity range, we first filter out images according to the initial datasets' file structure, to exclude pairings of different views of the same item, thus enhancing the diversity of the CIR triplets.
Next, we encode each image using the CLIP image encoder~\cite{radford2021clip}, and compute its cosine similarity with all other images in the same image source and category. 
Inspired by \cite{liu2021cirr}, we randomly select one image among the top-20 most similar images to form the image pair. 
This randomized selection enhances dataset diversity, preventing consistent pairing with the most similar images.

\begin{table}[t]
\caption{Quality evaluation of Fashion IQ and our FACap
dataset. We randomly sample 216 triplets across categories for each dataset and ask three annotators to measure data quality from three aspects with scale from 1 (worst) to 5 (best).}
\label{tab:dataset_quality_eval}
\begin{tabular}{lccc} \toprule
 & Faithfulness & Details & Saliency \\ \midrule
Fashion IQ~\cite{wu2020fashioniq} & \textbf{4.48 $\pm$ 0.64} & 3.03 $\pm$ 0.67 & 3.60 $\pm$ 0.69 \\
FACap & 4.40 $\pm$ 0.60 & \textbf{4.09 $\pm$ 0.64} & \textbf{4.29 $\pm$ 0.60} \\
\bottomrule
\end{tabular}
\end{table}

\noindent\textbf{Two-stage annotation}
We aim to utilize large vision and language foundation models~\cite{li2022blip,li2023blip2,chen2024internvl} to automatically annotate image pairs. 
However, currently, only a few VLMs are capable of accurately comparing two images in detail, and they often struggle to directly generate accurate modification texts from two images due to the scarcity of multi-image training data, as observed in our initial experiments.
Therefore, we propose a two-stage approach to generate more accurate and detailed image pair annotations. 

In the first stage, we use the open-source VLM model InternVL~\cite{chen2024internvl}, due to its good performance and modest computational requirements. The maximum token length for generation is set to 128, allowing the creation of long captions that capture as many fine-grained details as possible. These detailed captions are key in enhancing the precision of the modification texts generated in the second stage. To improve captioning accuracy and mitigate hallucinations which could introduce wrong elements in the caption, we prompt InternVL with the image category and available product descriptions and attributes from the image source. Although these additional inputs may be noisy, they often provide valuable context. Processing the entire dataset takes about 41 GPU hours on Nvidia A100 GPUs.
In the second stage, we use the proprietary LLM GPT-4o mini \cite{achiam2023gpt} \footnote{\href{https://platform.openai.com/docs/models/gpt-4o-mini\#gpt-4o-mini}{https://platform.openai.com/docs/models/gpt-4o-mini\#gpt-4o-mini}} to synthesize modification texts, benefitting from GPT4's strong capabilities in text reasoning. To guide the model in generating short and concise modification texts, we use clear instructions along with two in-context examples. This ensures that the LLM focuses on the most significant changes between the reference and target images.
Figure~\ref{fig:facap_data_examples} shows examples from FACap across different fashion categories.

\subsection{Quality evaluation}
\label{sec:dataset_eval}

Given the large size of FACap, manually and exhaustively evaluating its quality is challenging.
Therefore, we randomly sample 216 triplets from the Fashion IQ and the same number from the FACap dataset to assess the quality of their modification texts.
We evaluate the modification texts based on three key aspects:
\begin{itemize}
    \item \textbf{Faithfulness}: Whether the modification text accurately describes the changes between the reference image and target images. Note that this criteria indirectly evaluates the presence of hallucinations generated by the VLM and LLM, as they lead to inaccurate differences.
    \item \textbf{Details}: Whether the modification text captures multiple elements present in the images.
    \item \textbf{Saliency}: Whether the modification text focuses on unique elements, reducing the number of possible false-positive target images. A vague text could have a high faithfulness value, but would score poorly on saliency.
\end{itemize}
Each criterion is manually scored by three annotators on a scale from 1 (worst) to 5 (best) for each randomly sampled triplet.
The results are presented in Table~\ref{tab:dataset_quality_eval}. Notably, compared to the manually annotated Fashion IQ dataset for CIR, our automatically constructed dataset exhibits even higher quality. The similar faithfulness values indicate that our pipeline's caption errors are comparable to the rate of mistakes made by human annotators, while improving the amount of details and the relevance of the texts for retrieval, as shown by the details and saliency values.
This demonstrates the effectiveness of our annotation pipeline.

\subsection{Dataset Statistics}
\label{sec:dataset_stats}

Table~\ref{tab:data_stats} compares our FACap dataset with existing caption datasets in both the general and fashion domains.
FACap offers two key advantages over existing CIR fashion datasets. First, it significantly expands the dataset's size in the fashion domain with minimal additional time and cost. 
The scale of FACap is closer to that of general-domain image-text datasets like MSCOCO~\cite{lin2014microsoft}, and we exclude other web sources to ensure our dataset can be publicly available.
Second, FACap includes more accurate and detailed captions than existing datasets, as evidenced by our quality evaluation and average caption length. This can benefit fashion CIR tasks for fine-grained understanding of image-text alignment, particularly for fashion-related features and modifications, as illustrated in our qualitative results in figure \ref{fig:qualitative_examples}.

\section{The FashionBLIP-2 Model}

\subsection{Overall Framework}
\label{sec:overview}

Given a reference image $I_r$ and modification text $T$, the objective of CIR is to retrieve the correct target image $I_t$ from an image database $\mathcal{D}$. The retrieved image $I_t$ should accurately reflect the specified modifications applied to $I_r$.

\begin{figure*}[t]
    \centering
    \includegraphics[width=\linewidth]{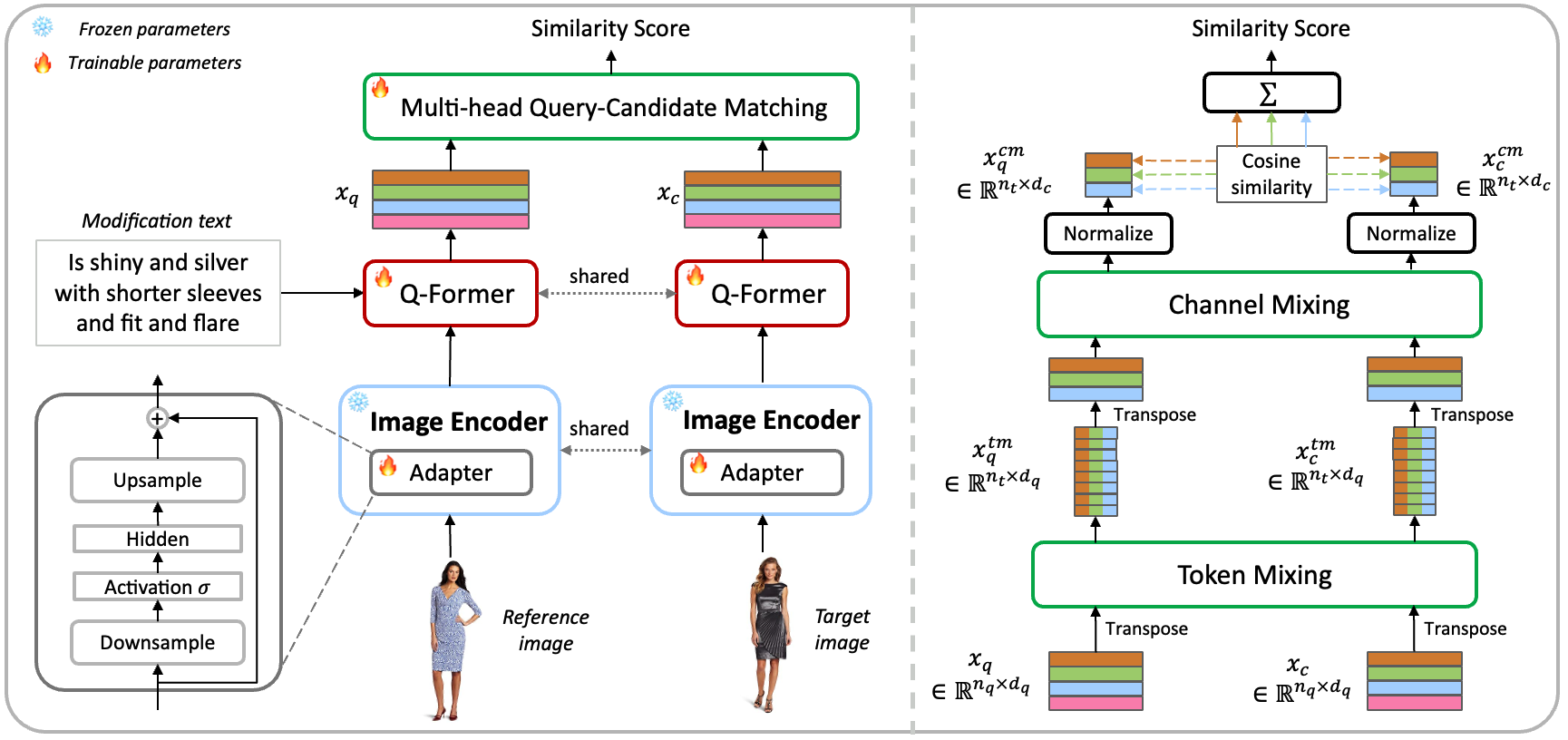}  
    \caption{Overview of the FashionBLIP-2 model. {\normalfont Left: The input images are encoded using a pretrained image encoder with adapter modules, and further processed by a Q-Former module. The similarity between the two obtained representations is computed using  multi-head query-candidate matching module. Right: Details of the matching module. The number of tokens and token dimensionality is reduced by token mixing and channel mixing respectively. The final similarity score is the sum of the cosine similarity for each paired vector.}}
    \label{fig:fashionblip2_architecture}
\end{figure*}

Figure~\ref{fig:fashionblip2_architecture} provides an overview of our FashionBLIP-2 model for the CIR task, which consists of three key modules: an image encoder for extracting image features, a light-weight Q-Former for compressing image features and performing multimodal fusion with text features, and a matching module for computing similarity between the query and the target image.
The image encoder and Q-Former are adapted from the pretrained BLIP-2 model~\cite{li2023blip2}, to which we refer readers for a more detailed explanation.

Given $I_r$, the image encoder first extracts a feature map $f_{r} \in \mathbb{R}^{h \times w \times d_I}$,  with $h, w$ the height and width of the encoded feature map and $d_I$ the feature dimensionality.
Then, the Q-Former employs a set of learnable queries to distill $f_{r}$ into a compact set of embeddings $x_{q} \in \mathbb{R}^{n_q \times d_q}$ together with guidance from the modification text $T$, where $n_q \ll h \times w$.
Similarly, each candidate image $I_c \in \mathcal{D}$ is sequentially processed by the image encoder and Q-Former but without any textual input, producing a corresponding set of embeddings $x_c \in \mathbb{R}^{n_q \times d_q}$ per image. 
Finally, the matching module takes the multimodal query embeddings $x_q$ and the candidate image embedding $x_c$ as inputs, computing a similarity score $s_{qc}$ between the query and the candidate image.
During inference, the similarity between the multimodal query and all candidate images is computed. The candidate images are then ranked in descending order based on their similarity scores, resulting in the final retrieval list.

\subsection{Adapter in Image Encoder}
\label{sec:adapter_module}

The BLIP-2 model~\cite{li2023blip2} is initially trained on large-scale open-domain datasets, potentially reducing its effectiveness at capturing fine-grained visual details crucial to the fashion domain, such as features related to sleeve length or specific collar types.
A straightforward approach to address this limitation is to fine-tune the BLIP-2 model alongside the CIR modules, to better adapt it to the fashion domain, but this may lead to high computational costs and catastrophic forgetting.

To overcome this challenge, we draw inspiration from~\cite{sung2022vl} and introduce lightweight adapter modules into each transformer layer~\cite{waswani2017attention} of the image encoder. Instead of fine-tuning the entire BLIP-2 backbone, we freeze the pretrained weights in the image encoder and train only the newly introduced adapter modules along with the lightweight Q-Former. 
As illustrated in Figure~\ref{fig:fashionblip2_architecture}, each adapter module comprises a downsampling layer, a non-linear operation, and an upsampling layer, with a residual link.
Given an input $x \in \mathbb{R}^c$: 
\begin{equation}
    \text{Adapter}(x) = x + W_u \left( \sigma( W_d x ) \right)
\end{equation}
where $W_d \in \mathbb{R}^{c_b \times c}$, $W_u \in \mathbb{R}^{c \times c_b}$ are trainable parameters with $c_b \ll c$, and $\sigma$ denotes the GELU function \cite{hendrycks2016gaussian}. 

This bottleneck architecture introduces a relatively small amount of additional parameters to the image encoder, ensuring a lightweight adaptation. 
Furthermore, the residual connection facilitates efficient gradient flow and helps preserve the original pretrained features, allowing the model to retain general-domain knowledge while learning fashion-specific details.

\subsection{Multi-head Query-Candidate Matching}
\label{sec:query_target_matching}

Previous works~\cite{liu2024bi,bai2024sentence} average the token embeddings $x_q$ and $x_c$ over the token dimension to obtain a single global vector for the query and each candidate image. However, this approach often suffers from the loss of fine-grained details crucial for retrieval.

To address this, we propose a multi-head query-candidate matching method based on a dual-level mixing operation like~\cite{tolstikhin2021mlp} to better capture fine-grained information.
First, we use token mixing over the input tokens for $x_q \in \mathbb{R}^{n_q \times d_q}$, formulated as:
\begin{equation}
    x^{tm}_q = W_{tm} \times x_q
\end{equation}
where $W_{tm} \in \mathbb{R}^{n_t \times n_q}$ is a trainable parameter.
Here, $n_t < n_q$ to reduce the redundancy across embeddings in the initial representation $x_q$, while retaining multiple vectors to capture multiple aspects of the inputs.
We then perform channel mixing for each vector to project $x^{tm}_q$ into a lower-dimensional space:
\begin{equation}
    x^{cm}_q = x_q \times W_{cm}
\end{equation}
where $W_{cm} \in \mathbb{R}^{d_q \times d_c}$ with $d_c < d_q$.
In order to encourage projecting $x_q$ and $x_c$ into a common low-dimensional embedding space, we use the same parameters $W_{tm} = W_{cm}$ to process $x_q$ and $x_c$.

Each row vector in $x^{cm}_q$ and $x^{cm}_c$ is viewed as one head for matching. The final similarity is the sum of the cosine similarities for each head as follows:
\begin{equation}
   s_{qc} = \text{sim}\left(x^{cm}_q, x^{cm}_c \right) = \sum_{i=1}^{n_t} \frac{x^{cm}_{q,i} \cdot x^{cm}_{c, i}} {\lVert x^{cm}_{q,i} \rVert_2 \cdot \lVert x^{cm}_{c,i} \rVert_2}
\end{equation}

\subsection{Training}
\label{sec:training}

We train the FashionBLIP-2 model in two stages. 

\noindent\textbf{Stage 1: Training on FACap.}
The first stage aims to fine-tune a general-domain model to the fashion domain, to learn fine-grained visual and text representations.
Since FACap contains both CIR triplets and image-caption pairs, we use two tasks in stage 1 training: the primary CIR task and an auxiliary Composed Text Retrieval (CTR) task.

For the CIR task, we employ the widely-used contrastive loss:
\begin{equation}
\label{eq:cir_loss}
\mathcal{L}_{\text{CIR}} = - \frac{1}{n} \sum_{i=1}^n \log \left(     
\frac{\exp(x_i y_i)}{\exp(x_i y_i) + \sum \limits_{\hat{y} \in \mathcal{N}_i} \exp(x_i \hat{y})} \right)
\end{equation}
with $(x_i, y_i)$ a positive pair, and $\mathcal{N}_i$ the set of negative pairs. Here, the negative pairs correspond to the reference image $x_i$ and any target image other than $y_i$ in the batch. 

The CTR task retrieves a target text corresponding to the target image from a text pool, rather than the target image as in CIR. This auxiliary task helps the model align the fused query embedding more effectively with the textual representation, complementing its alignment with the image representation in CIR task. The contrastive loss used for CTR task, $\mathcal{L}_{\text{CTR}}$, is defined as in Eq~\ref{eq:cir_loss}.

We fine-tune the whole FashionBLIP-2 except for the original image encoder, using the combined loss function $\mathcal{L}_{\text{CIR}} + \mathcal{L}_{\text{CTR}}$.

\noindent\textbf{Stage 2: Fine-tuning on downstream Fashion CIR dataset.}
The second stage fine-tunes the FashionBLIP-2 model on the downstream dataset to further improve its performance. However, since Fashion IQ does not contain image-caption pairs, we train the model exclusively with $\mathcal{L}_{\text{CIR}}$. Additionally, we freeze the image encoder and its adapter modules to preserve the fashion-domain knowledge learnt during the first stage of training.

\section{Experiment}

\begin{table*}
  \caption{Results on the Fashion IQ validation split for composed image retrieval, under two settings: with and without fine-tuning the model on Fashion IQ. Best and second-best results in each setting are highlighted in bold and underlined, respectively.}
  \label{tab:evaluation_fashioniq}
  \begin{tabular}{c|c|cc|cc|cc|c}
    \toprule
    \multirow{2}{*}{\textbf{Setting}} & \multirow{2}{*}{\textbf{Model}} & \multicolumn{2}{c}{\textbf{Dresses}} & \multicolumn{2}{c}{\textbf{Shirts}} & \multicolumn{2}{c}{\textbf{Tops\&tees}} & \multirow{2}{*}{\textbf{Average}} \\
      & & R@10 & R@50 & R@10 & R@50 & R@10 & R@50 & \\
    \midrule
    \multirow{6}{*}{\makecell{Without \\ fine-tuning}} & Pic2Word \cite{saito2023pic2word} & 20.00 & 40.20 & 26.20 & 43.60 & 27.90 & 47.40 & 34.22 \\
      & SEARLE (ViT-L/14) \cite{baldrati2023zero} & 20.48 & 43.13 & 26.89 & 45.58 & 29.32 & 49.97 & 35.90 \\
      & Context-I2W \cite{tang2024context} & 23.1 & 45.3 & 29.7 & 48.6 & 30.6 & 52.9 & 39.37 \\
      & FTI4CIR \cite{lin2024fine} & 24.39 & 47.84 & 31.35 & 50.59 & 32.43 & 54.21 & 40.14 \\
      & LDRE (ViT-G/14) \cite{yang2024ldre} & \underline{26.11} & \underline{51.12} & \textbf{35.94} & \textbf{58.58} & \underline{35.42} & \underline{56.67} & \underline{43.97} \\ \cmidrule{2-9}
      & FashionBLIP-2 (ours) & \textbf{32.52} & \textbf{53.25} & \underline{34.79} & \underline{52.40} & \textbf{36.66} & \textbf{58.13} & \textbf{44.63} \\
    
    \midrule\midrule

    \multirow{9}{*}{\makecell{With \\ fine-tuning}} & CLIP4CIR~\cite{baldrati2022conditioned} & 33.81 & 59.40 & 39.99 & 60.45 & 41.41 & 65.37 & 50.03 \\
      & BLIP4CIR+Bi~\cite{liu2024bi} & 42.09 & 67.33 & 41.76 & 64.28 & 46.61 & 70.32 & 55.40 \\
      & BLIP2-Cir~\cite{jiang2024cala} & 41.57 & 66.02 & 46.86 & 66.00 & 49.44 & 72.25 & 57.02 \\
      & TG-CIR~\cite{wen2023target} & 45.22 & 69.66 & 52.60 & 72.52 & 56.14 & 77.10 & 58.05 \\
     & FAME-ViL~\cite{han2023fame} & 42.19 & 67.38 & 47.64 & 68.79 & 50.69 & 73.07 & 58.29 \\
      & Re-ranking~\cite{liu2024candidate} & 48.14 & 71.43 & 50.15 & 71.25 & 55.23 & 76.80 & 62.15 \\
      & SPRC~\cite{bai2024sentence} & 49.18 & 72.43 & 55.64 & 73.89 & \underline{59.35} & 78.58 & 64.85 \\
      & UniFashion~\cite{zhao2024unifashion} & \textbf{53.72} & \textbf{73.66} & \textbf{61.25} & \textbf{76.67} & \textbf{61.84} & \textbf{80.46} & \textbf{67.93} \\ \cmidrule{2-9}
      & FashionBLIP-2 (ours) & \underline{51.41} & \underline{73.53} & \underline{57.02} & \underline{75.32} & 58.95 & \underline{79.60} & \underline{65.97} \\ 
    \bottomrule
  \end{tabular}
\end{table*}

\subsection{Experimental setup}

\noindent\textbf{Datasets.}
We use the Fashion IQ dataset~\cite{wu2020fashioniq} for evaluation, which is the most widely used Fashion CIR dataset. With 18,000 training triplets and 6,016 validation triplets, it covers three categories: Dress, Shirt, and Toptee. Each reference and target image pair contains two manually annotated modification texts. Following previous works~\cite{baldrati2022conditioned,liu2024bi,feng2024improving,liu2024candidate,zhao2022progressive,zhao2024unifashion}, we concatenate the two annotated texts with an ``and'' word to form a single modification text, and evaluate the models on the validation split.

However, since the annotations in Fashion IQ are noisy as shown in Figure~\ref{fig:teaser} and Table~\ref{tab:dataset_quality_eval}, we enhance its quality by applying our automatic annotation process to the images from the Fashion IQ validation split.
We create a triplet for each unique image in the validation split and generate a total of 15,536 CIR triplets, which we name enhFashionIQ, for fine-grained CIR evaluation.

\noindent\textbf{Evaluation metrics.}
We use Recall@$k$ (with $k \in \{10; 50\}$ similarly to previous works) as the main metric. It computes the percentage of target images that appear in the top-$k$ retrieved images list. The recalls are computed for each category: dress, shirt, and toptee for Fashion IQ and enhFashionIQ. We also report the average recall.

\newpage

\noindent\textbf{Experiment settings.}
We evaluate models under two settings:
\begin{itemize}
    \item without fine-tuning setting: the model is only trained on our FACap dataset and then evaluated on downstream CIR datasets. This setting evaluates the generalization capacity of the model on a previously unseen fashion dataset, and its performance is compared to zero-shot methods~\cite{baldrati2023zero, saito2023pic2word, tang2024context, yang2024ldre, lin2024fine, karthik2023vision}.
    \item fine-tuning setting: the model is  fine-tuned on Fashion IQ, and evaluated on Fashion IQ and enhFashionIQ. 
\end{itemize}

\noindent\textbf{Implementation details}
We use the ViT-G version of the pre-trained BLIP-2 \cite{li2023blip2} model. For the adapter module in image encoder, we use a downsampling factor of 16. 
The Q-Former module is parametrized to take textual inputs of 128 tokens and $n_q = 32$ query tokens, and outputs 32 token embeddings with dimensionality of $d_q = 768$.
The token mixing layer in our multi-head query-target matching reduces the 32 tokens to $n_t = 12$ tokens and channel dimension from 768 to $d_c = 256$. 
We run our experiments on NVIDIA H100 GPUs with batch size of 512 and AdamW optimizer.

\subsection{Comparison with state-of-the-art methods}

Table~\ref{tab:evaluation_fashioniq} presents the evaluation results on Fashion IQ under the two settings - without and with fine-tuning.
In the upper block, we compare the FashionBLIP-2 model, trained only on our FACap dataset, with zero-shot methods~\cite{saito2023pic2word,baldrati2023zero,tang2024context,yang2024ldre,lin2024fine} to compare their generalization capacity on fashion data. 
Our model achieves an improvement over the state-of-the-art method LDRE~\cite{lin2024fine}, which uses pre-trained LLMs.
The average gain is 0.66 absolute points with 2.17 in R@10, highlighting its ability to retrieve relevant images on a previously unseen fashion dataset.
The most pronounced improvement is observed in the dress category, known for its high diversity in descriptions such as their length, pattern, neckline, and way of wearing, further demonstrating the effectiveness of our proposed dataset and approach.

The bottom section of Table~\ref{tab:evaluation_fashioniq} shows the comparison between our FashionBLIP-2 and existing methods~\cite{baldrati2022conditioned,liu2024bi,jiang2024cala,wen2023target,han2023fame,liu2024candidate,bai2024sentence,zhao2024unifashion} fine-tuned on the Fashion IQ dataset.
Our FashionBLIP-2 achieves the second best results on average, only under-performing UniFashion~\cite{zhao2024unifashion} which utilizes more image-caption pairs \textemdash about 280k pairs \textemdash and generation tasks in training rather than CIR triplets. 
As our FACap dataset and method are complementary to UniFashion, we will leave it to future work.

\subsection{Ablation study}

\noindent\textbf{Pretraining on the FACap dataset.}
In  Table~\ref{tab:ablation_fashionblip2_vs_sprc}, we evaluate the contribution of the proposed FACap dataset using our FashionBLIP-2 model, SPRC~\cite{bai2024sentence} and UniFashion~\cite{zhao2024unifashion}.
The evaluation is conducted on both Fashion IQ and our enhFashionIQ, containing more fine-grained annotations.
First, almost all models benefit from pretraining on FACap, improving performance on the two evaluation datasets. 
Second, our FACap dataset provides greater benefits to the FashionBLIP-2 model. 
This advantage is attributed to our model's multi-head matching mechanism, which effectively leverages the fine-grained details in FACap, whereas SPRC struggles to utilize such detailed information due to its reliance on global embeddings.
Finally, FashionBLIP-2 achieves a better performance than SPRC on the Fashion IQ split, and it demonstrates significantly higher improvements on enhFashionIQ under the same training configuration. This highlights the superior ability of our model to handle fine-grained fashion retrieval tasks.
In addition, we observe that fine-tuning our model on Fashion IQ does not degrade much its performance on enhFashionIQ compared to SPRC. This indicates that our method is more robust to noisy datasets, without losing its fine-grained performance.

\begin{table*}
  \caption{Impact of FACap pretraining on SPRC~\cite{bai2024sentence}, UniFashion~\cite{zhao2024unifashion} (code reproduction) and our method. 
  The averaged Recall is reported for Fashion IQ and enhFashionIQ validation splits.}
  \label{tab:ablation_fashionblip2_vs_sprc}
  \begin{threeparttable}
  \begin{tabular}{c|cccc}
    \toprule
    \multirow{2}{*}{\textbf{Model}} & \multirow{2}{*}{\makecell{\textbf{Pretrain} \\ \textbf{on FACap}}} & \multirow{2}{*}{\makecell{\textbf{Fine-tune} \\ \textbf{on Fashion IQ}}} & \multirow{2}{*}{\textbf{Fashion IQ}} & \multirow{2}{*}{\textbf{enhFashionIQ}} \\
    &  &  &  &  \\
    \midrule
    \multirow{2}{*}{SPRC~\cite{bai2024sentence}} & \xmark & \cmark & 64.85 & 79.59 \\ 
     & \cmark & \cmark & 64.87 & 80.29 \\ 
    \midrule
    \multirow{2}{*}{UniFashion~\cite{zhao2024unifashion}$^*$} & \xmark & \cmark & \underline{65.34} & 81.97 \\
     & \cmark & \cmark & 64.51 & \underline{87.30} \\
    \midrule
    \multirow{2}{*}{\begin{tabular}[c]{@{}c@{}}FashionBLIP-2\\ (Ours)\end{tabular}} & \xmark & \cmark & 64.46 & 80.32 \\
     & \cmark & \cmark & \textbf{65.97} & \textbf{87.93} \\
    
    \bottomrule
  \end{tabular}
  \begin{tablenotes}
    \item[*] Results obtained using UniFashion's released code
  \end{tablenotes}
  \end{threeparttable}
\end{table*}

\begin{table}[t]
    \caption{Ablation study of different components in FashionBLIP-2 model.
    The averaged Recall is reported for Fashion IQ and enhFashionIQ validation splits under two settings. The acronym ``MH" denotes the proposed multi-head matching.}
    \label{tab:ablation_component}
    \begin{tabular}{ccccccc} \toprule
    \multirow{2}{*}{\begin{tabular}[c]{@{}c@{}}\textbf{Training}\\\textbf{tasks}\end{tabular}} & \multirow{2}{*}{\textbf{Adapter}} & \multirow{2}{*}{\textbf{Matching}} & \multicolumn{2}{c}{\textbf{no fine-tuning}} & \multicolumn{2}{c}{\textbf{w/ fine-tuning}} \\
    &  &  & Fashion IQ & enhFashionIQ & Fashion IQ & enhFashionIQ \\ \midrule
    CIR & \xmark & Global & 41.04 & 87.81 & 64.36 & 86.42 \\
    CIR & \cmark & Global & 42.62 & 88.20 & 64.38 & 86.75 \\
    CIR & \cmark & MH & 43.31 & 89.14 & \textbf{65.97} & \textbf{87.93} \\
    CIR+CTR & \cmark & MH & \textbf{44.63} & \textbf{89.45} & 65.62 & 86.93\\
    \bottomrule
    \end{tabular}
\end{table}

\noindent\textbf{FashionBLIP-2 components.}
Table~\ref{tab:ablation_component} analyzes the individual contributions of each component in our FashionBLIP-2 model.
The first row serves as the baseline, representing a model built on top of BLIP-2. 
In the second row, we specialize the image encoder with adapter modules, improving performance across both datasets and evaluation settings.
The third row incorporates the proposed multi-head query-candidate matching mechanism. This boosts retrieval performance by enabling the model to capture and compare finer details between queries and candidates.
Finally, the fourth row integrates the auxiliary CTR task during training on the FACap dataset. While it improves results in the setting without fine-tuning on Fashion IQ, it decreases performance when fine-tuned on Fashion IQ. We hypothesize that the CTR task may introduce a bias towards detailed textual descriptions, which might hinder adaptation to noisier datasets like Fashion IQ.

\begin{figure}
    \centering
    \includegraphics[width=0.8\linewidth]{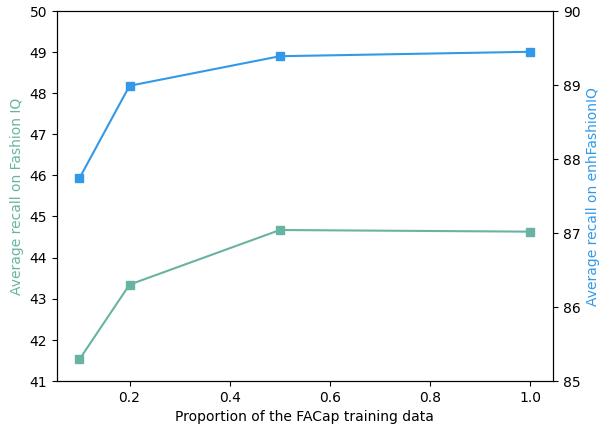}
    \caption{Results on Fashion IQ dataset without fine-tuning using different sizes of the FACap dataset.}
    \label{fig:ablation_study_datasize}
\end{figure}

\noindent\textbf{Size of the training data.}
To investigate the impact of data quantity, we train FashionBLIP-2 on progressively larger subsets of FACap and evaluate the resulting models on Fashion IQ and enhFashionIQ.
As shown in Figure~\ref{fig:ablation_study_datasize}, the performance of our model improves with the increasing size of the training dataset across both evaluation benchmarks. This highlights the critical role of having a large volume of diverse image-text pairs to effectively learn fine-grained multimodal representations.
However, the performance gain from training on 50\% to 100\% of the dataset is relatively small: while data quantity is important, further improvements may require focusing on data quality and diversity rather than sheer volume.

\begin{figure*}[t]
    \centering
    \includegraphics[width=0.8\linewidth]{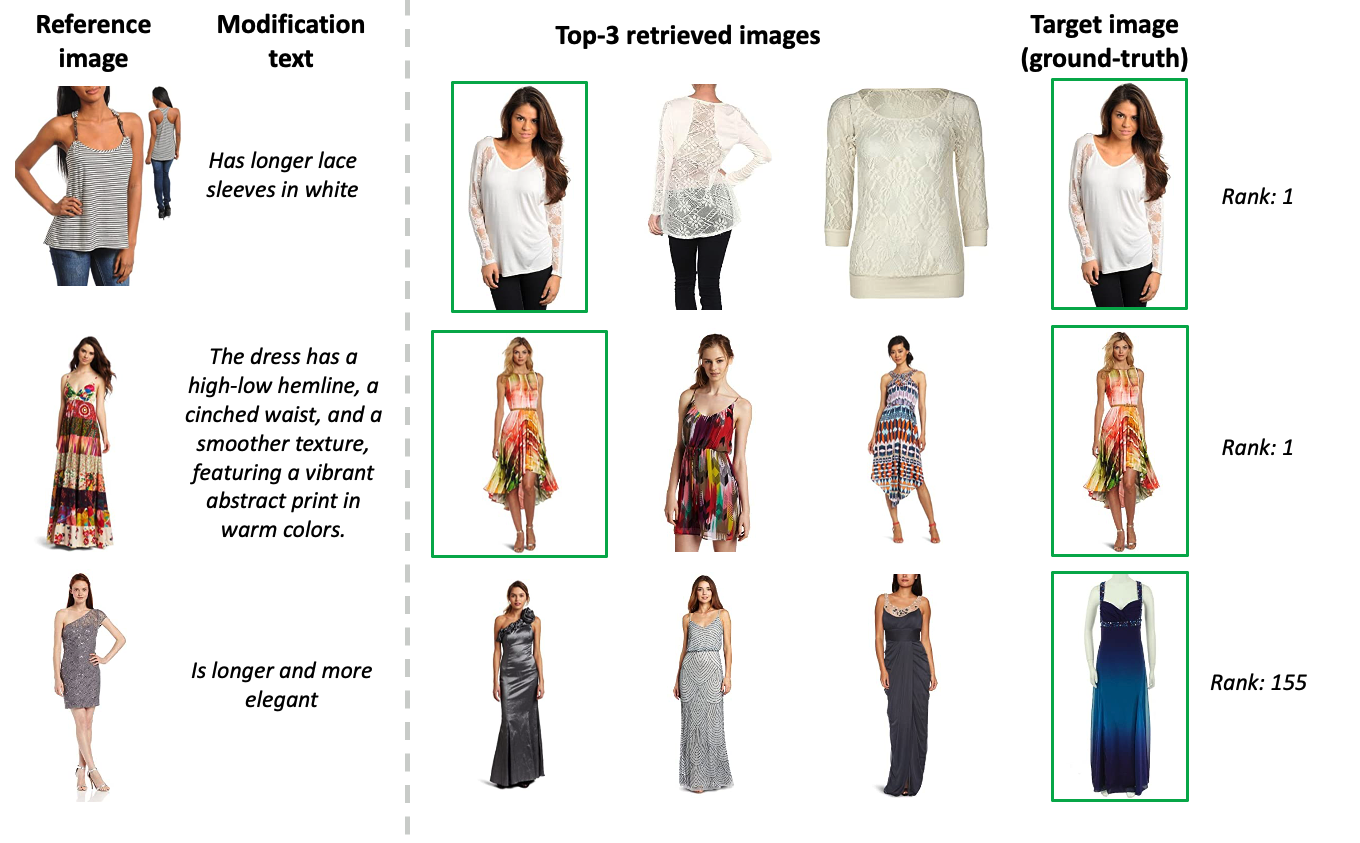}  
    \caption{Qualitative results of FashionBLIP-2 on Fashion IQ (rows 1 and 3) and enhFashionIQ (row 2). The rank of the ground-truth image (framed in green) among the retrieved results is specified on the right.}
    \label{fig:qualitative_examples}
\end{figure*}

\subsection{Qualitative Results}
Figure \ref{fig:qualitative_examples} presents qualitative results of FashionBLIP-2 on the Fashion IQ and enhFashionIQ validation data. The first two rows show that whether the modification text is precise (enhFashionIQ) or not (Fashion IQ), our model is able to combine it with characteristics of the reference images (for example clothing length and color). 
The third row presents a failure case of our model, revealing the difficulty of handling false negative examples: the correct target image is badly ranked, but all the top-3 retrieved images respect the given modification text and the information from the reference image (color and sleeves).

\section{Conclusion}
We have proposed two enhancements to tackle shortcomings of CIR in the fashion domain. 
Firstly, we designed an automatic pipeline to build a large-scale high-quality CIR dataset from a large list of images with noisy captions. Leveraging the strengths of a VLM and a LLM, this pairing and annotation method provides accurate modification texts, while adding relevant fashion details and focusing on salient changes. This method has allowed us to construct FACap, a higher quality dataset for fashion CIR. 
Secondly, we have introduced FashionBLIP-2, a method combining BLIP-2's general-domain comprehensive strength with an adapter module to adjust it to the fashion domain, and a new multi-head query-candidate matching mechanism to focus on fine-grained details and benefit from FACap high-quality captioning.
Experiments show that FashionBLIP-2  trained on FACap outperforms state-of-the-art methods without fine-tuning on the downstream dataset. It also reaches competitive performance after fine-tuning on Fashion IQ, making it well-suited for fast adaptation in the fashion domain, excelling in fine-grained retrieval tasks while remaining robust to vague modification texts.

\begin{acknowledgments}
    This project was granted access to the HPC resources of IDRIS under the allocation AD011015247R1 made by GENCI. It was funded in part by the French government under management of Agence Nationale de la Recherche as part of the "France 2030" program, reference ANR-23-IACL-0008 (PR[AI]RIE-PSAI project), and Paris Île-de-France Région in the frame of the DIM AI4IDF.
\end{acknowledgments}

\section*{Declaration on Generative AI}
  The authors have not employed any Generative AI tools for the writing of this paper.
  
\bibliography{sample-ceur}

\begin{thebibliography}{56}
\expandafter\ifx\csname natexlab\endcsname\relax\def\natexlab#1{#1}\fi
\providecommand{\url}[1]{\texttt{#1}}
\providecommand{\href}[2]{#2}
\providecommand{\path}[1]{#1}
\providecommand{\DOIprefix}{doi:}
\providecommand{\ArXivprefix}{arXiv:}
\providecommand{\URLprefix}{URL: }
\providecommand{\Pubmedprefix}{pmid:}
\providecommand{\doi}[1]{\href{http://dx.doi.org/#1}{\path{#1}}}
\providecommand{\Pubmed}[1]{\href{pmid:#1}{\path{#1}}}
\providecommand{\bibinfo}[2]{#2}
\ifx\xfnm\relax \def\xfnm[#1]{\unskip,\space#1}\fi
\bibitem[{Wu et~al.(2021)Wu, Gao, Guo, Al-Halah, Rennie, Grauman, and Feris}]{wu2020fashioniq}
\bibinfo{author}{H.~Wu}, \bibinfo{author}{Y.~Gao}, \bibinfo{author}{X.~Guo}, \bibinfo{author}{Z.~Al-Halah}, \bibinfo{author}{S.~Rennie}, \bibinfo{author}{K.~Grauman}, \bibinfo{author}{R.~Feris},
\newblock \bibinfo{title}{Fashion iq: A new dataset towards retrieving images by natural language feedback},
\newblock in: \bibinfo{booktitle}{Proceedings of the IEEE/CVF Conference on computer vision and pattern recognition}, \bibinfo{year}{2021}, pp. \bibinfo{pages}{11307--11317}.
\bibitem[{Shao et~al.(2023)Shao, Chen, Karpur, Cui, Araujo, and Cao}]{shao2023global}
\bibinfo{author}{S.~Shao}, \bibinfo{author}{K.~Chen}, \bibinfo{author}{A.~Karpur}, \bibinfo{author}{Q.~Cui}, \bibinfo{author}{A.~Araujo}, \bibinfo{author}{B.~Cao},
\newblock \bibinfo{title}{Global features are all you need for image retrieval and reranking},
\newblock in: \bibinfo{booktitle}{Proceedings of the IEEE/CVF International Conference on Computer Vision}, \bibinfo{year}{2023}, pp. \bibinfo{pages}{11036--11046}.
\bibitem[{Rao et~al.(2022)Rao, Wang, Ding, Qi, Zhan, Liu, and Tao}]{rao2022does}
\bibinfo{author}{J.~Rao}, \bibinfo{author}{F.~Wang}, \bibinfo{author}{L.~Ding}, \bibinfo{author}{S.~Qi}, \bibinfo{author}{Y.~Zhan}, \bibinfo{author}{W.~Liu}, \bibinfo{author}{D.~Tao},
\newblock \bibinfo{title}{Where does the performance improvement come from? -a reproducibility concern about image-text retrieval},
\newblock in: \bibinfo{booktitle}{Proceedings of the 45th international ACM SIGIR conference on research and development in information retrieval}, \bibinfo{year}{2022}, pp. \bibinfo{pages}{2727--2737}.
\bibitem[{Vo et~al.(2019)Vo, Jiang, Sun, Murphy, Li, Fei-Fei, and Hays}]{vo2019composing}
\bibinfo{author}{N.~Vo}, \bibinfo{author}{L.~Jiang}, \bibinfo{author}{C.~Sun}, \bibinfo{author}{K.~Murphy}, \bibinfo{author}{L.-J. Li}, \bibinfo{author}{L.~Fei-Fei}, \bibinfo{author}{J.~Hays},
\newblock \bibinfo{title}{Composing text and image for image retrieval-an empirical odyssey},
\newblock in: \bibinfo{booktitle}{Proceedings of the IEEE/CVF conference on computer vision and pattern recognition}, \bibinfo{year}{2019}, pp. \bibinfo{pages}{6439--6448}.
\bibitem[{Baldrati et~al.(2022)Baldrati, Bertini, Uricchio, and Del~Bimbo}]{baldrati2022conditioned}
\bibinfo{author}{A.~Baldrati}, \bibinfo{author}{M.~Bertini}, \bibinfo{author}{T.~Uricchio}, \bibinfo{author}{A.~Del~Bimbo},
\newblock \bibinfo{title}{Conditioned and composed image retrieval combining and partially fine-tuning clip-based features},
\newblock in: \bibinfo{booktitle}{Proceedings of the IEEE/CVF Conference on Computer Vision and Pattern Recognition}, \bibinfo{year}{2022}, pp. \bibinfo{pages}{4959--4968}.
\bibitem[{Liu et~al.(2024)Liu, Sun, Hong, Teney, and Gould}]{liu2024bi}
\bibinfo{author}{Z.~Liu}, \bibinfo{author}{W.~Sun}, \bibinfo{author}{Y.~Hong}, \bibinfo{author}{D.~Teney}, \bibinfo{author}{S.~Gould},
\newblock \bibinfo{title}{Bi-directional training for composed image retrieval via text prompt learning},
\newblock in: \bibinfo{booktitle}{Proceedings of the IEEE/CVF Winter Conference on Applications of Computer Vision}, \bibinfo{year}{2024}, pp. \bibinfo{pages}{5753--5762}.
\bibitem[{Feng et~al.(2024)Feng, Zhang, and Nie}]{feng2024improving}
\bibinfo{author}{Z.~Feng}, \bibinfo{author}{R.~Zhang}, \bibinfo{author}{Z.~Nie},
\newblock \bibinfo{title}{Improving composed image retrieval via contrastive learning with scaling positives and negatives},
\newblock in: \bibinfo{booktitle}{Proceedings of the 32nd ACM International Conference on Multimedia}, \bibinfo{year}{2024}, pp. \bibinfo{pages}{1632--1641}.
\bibitem[{Liu et~al.(2024)Liu, Sun, Teney, and Gould}]{liu2024candidate}
\bibinfo{author}{Z.~Liu}, \bibinfo{author}{W.~Sun}, \bibinfo{author}{D.~Teney}, \bibinfo{author}{S.~Gould},
\newblock \bibinfo{title}{Candidate set re-ranking for composed image retrieval with dual multi-modal encoder},
\newblock \bibinfo{journal}{Transactions on Machine Learning Research}  (\bibinfo{year}{2024}). \URLprefix \url{https://openreview.net/forum?id=fJAwemcvpL}.
\bibitem[{Zhao et~al.(2022)Zhao, Song, and Jin}]{zhao2022progressive}
\bibinfo{author}{Y.~Zhao}, \bibinfo{author}{Y.~Song}, \bibinfo{author}{Q.~Jin},
\newblock \bibinfo{title}{Progressive learning for image retrieval with hybrid-modality queries},
\newblock in: \bibinfo{booktitle}{Proceedings of the 45th International ACM SIGIR Conference on Research and Development in Information Retrieval}, \bibinfo{year}{2022}, pp. \bibinfo{pages}{1012--1021}.
\bibitem[{Zhao et~al.(2024)Zhao, Zhang, Zhang, and Wu}]{zhao2024unifashion}
\bibinfo{author}{X.~Zhao}, \bibinfo{author}{Y.~Zhang}, \bibinfo{author}{W.~Zhang}, \bibinfo{author}{X.-M. Wu},
\newblock \bibinfo{title}{{U}ni{F}ashion: A unified vision-language model for multimodal fashion retrieval and generation},
\newblock in: \bibinfo{editor}{Y.~Al-Onaizan}, \bibinfo{editor}{M.~Bansal}, \bibinfo{editor}{Y.-N. Chen} (Eds.), \bibinfo{booktitle}{Proceedings of the 2024 Conference on Empirical Methods in Natural Language Processing}, \bibinfo{publisher}{Association for Computational Linguistics}, \bibinfo{address}{Miami, Florida, USA}, \bibinfo{year}{2024}, pp. \bibinfo{pages}{1490--1507}. \URLprefix \url{https://aclanthology.org/2024.emnlp-main.89/}. \DOIprefix\doi{10.18653/v1/2024.emnlp-main.89}.
\bibitem[{Radford et~al.(2021)Radford, Kim, Hallacy, Ramesh, Goh, Agarwal, Sastry, Askell, Mishkin, Clark et~al.}]{radford2021clip}
\bibinfo{author}{A.~Radford}, \bibinfo{author}{J.~W. Kim}, \bibinfo{author}{C.~Hallacy}, \bibinfo{author}{A.~Ramesh}, \bibinfo{author}{G.~Goh}, \bibinfo{author}{S.~Agarwal}, \bibinfo{author}{G.~Sastry}, \bibinfo{author}{A.~Askell}, \bibinfo{author}{P.~Mishkin}, \bibinfo{author}{J.~Clark}, et~al.,
\newblock \bibinfo{title}{Learning transferable visual models from natural language supervision},
\newblock in: \bibinfo{booktitle}{International conference on machine learning}, \bibinfo{organization}{PMLR}, \bibinfo{year}{2021}, pp. \bibinfo{pages}{8748--8763}.
\bibitem[{Li et~al.(2023)Li, Li, Savarese, and Hoi}]{li2023blip2}
\bibinfo{author}{J.~Li}, \bibinfo{author}{D.~Li}, \bibinfo{author}{S.~Savarese}, \bibinfo{author}{S.~Hoi},
\newblock \bibinfo{title}{Blip-2: Bootstrapping language-image pre-training with frozen image encoders and large language models},
\newblock in: \bibinfo{booktitle}{International conference on machine learning}, \bibinfo{organization}{PMLR}, \bibinfo{year}{2023}, pp. \bibinfo{pages}{19730--19742}.
\bibitem[{Park et~al.(2019)Park, Darrell, and Rohrbach}]{park2019robust}
\bibinfo{author}{D.~H. Park}, \bibinfo{author}{T.~Darrell}, \bibinfo{author}{A.~Rohrbach},
\newblock \bibinfo{title}{Robust change captioning},
\newblock in: \bibinfo{booktitle}{Proceedings of the IEEE/CVF International Conference on Computer Vision}, \bibinfo{year}{2019}, pp. \bibinfo{pages}{4624--4633}.
\bibitem[{Baldrati et~al.(2023)Baldrati, Agnolucci, Bertini, and Del~Bimbo}]{baldrati2023zero}
\bibinfo{author}{A.~Baldrati}, \bibinfo{author}{L.~Agnolucci}, \bibinfo{author}{M.~Bertini}, \bibinfo{author}{A.~Del~Bimbo},
\newblock \bibinfo{title}{Zero-shot composed image retrieval with textual inversion},
\newblock in: \bibinfo{booktitle}{Proceedings of the IEEE/CVF International Conference on Computer Vision}, \bibinfo{year}{2023}, pp. \bibinfo{pages}{15338--15347}.
\bibitem[{Saito et~al.(2023)Saito, Sohn, Zhang, Li, Lee, Saenko, and Pfister}]{saito2023pic2word}
\bibinfo{author}{K.~Saito}, \bibinfo{author}{K.~Sohn}, \bibinfo{author}{X.~Zhang}, \bibinfo{author}{C.-L. Li}, \bibinfo{author}{C.-Y. Lee}, \bibinfo{author}{K.~Saenko}, \bibinfo{author}{T.~Pfister},
\newblock \bibinfo{title}{Pic2word: Mapping pictures to words for zero-shot composed image retrieval},
\newblock in: \bibinfo{booktitle}{Proceedings of the IEEE/CVF Conference on Computer Vision and Pattern Recognition}, \bibinfo{year}{2023}, pp. \bibinfo{pages}{19305--19314}.
\bibitem[{Tang et~al.(2024)Tang, Yu, Gai, Zhuang, Xiong, Hu, and Wu}]{tang2024context}
\bibinfo{author}{Y.~Tang}, \bibinfo{author}{J.~Yu}, \bibinfo{author}{K.~Gai}, \bibinfo{author}{J.~Zhuang}, \bibinfo{author}{G.~Xiong}, \bibinfo{author}{Y.~Hu}, \bibinfo{author}{Q.~Wu},
\newblock \bibinfo{title}{Context-i2w: Mapping images to context-dependent words for accurate zero-shot composed image retrieval},
\newblock in: \bibinfo{booktitle}{Proceedings of the AAAI Conference on Artificial Intelligence}, volume~\bibinfo{volume}{38}, \bibinfo{year}{2024}, pp. \bibinfo{pages}{5180--5188}.
\bibitem[{Yang et~al.(2024)Yang, Xue, Qian, Dong, and Xu}]{yang2024ldre}
\bibinfo{author}{Z.~Yang}, \bibinfo{author}{D.~Xue}, \bibinfo{author}{S.~Qian}, \bibinfo{author}{W.~Dong}, \bibinfo{author}{C.~Xu},
\newblock \bibinfo{title}{Ldre: Llm-based divergent reasoning and ensemble for zero-shot composed image retrieval},
\newblock in: \bibinfo{booktitle}{Proceedings of the 47th International ACM SIGIR Conference on Research and Development in Information Retrieval}, \bibinfo{year}{2024}, pp. \bibinfo{pages}{80--90}.
\bibitem[{Lin et~al.(2024)Lin, Wen, Song, Liu, Hu, and Nie}]{lin2024fine}
\bibinfo{author}{H.~Lin}, \bibinfo{author}{H.~Wen}, \bibinfo{author}{X.~Song}, \bibinfo{author}{M.~Liu}, \bibinfo{author}{Y.~Hu}, \bibinfo{author}{L.~Nie},
\newblock \bibinfo{title}{Fine-grained textual inversion network for zero-shot composed image retrieval},
\newblock in: \bibinfo{booktitle}{Proceedings of the 47th International ACM SIGIR Conference on Research and Development in Information Retrieval}, \bibinfo{year}{2024}, pp. \bibinfo{pages}{240--250}.
\bibitem[{Karthik et~al.(2023)Karthik, Roth, Mancini, and Akata}]{karthik2023vision}
\bibinfo{author}{S.~Karthik}, \bibinfo{author}{K.~Roth}, \bibinfo{author}{M.~Mancini}, \bibinfo{author}{Z.~Akata},
\newblock \bibinfo{title}{Vision-by-language for training-free compositional image retrieval},
\newblock \bibinfo{journal}{arXiv preprint arXiv:2310.09291}  (\bibinfo{year}{2023}).
\bibitem[{Bai et~al.(2024)Bai, Xu, Liu, Khan, Khan, Zuo, Goh, Feng et~al.}]{bai2024sentence}
\bibinfo{author}{Y.~Bai}, \bibinfo{author}{X.~Xu}, \bibinfo{author}{Y.~Liu}, \bibinfo{author}{S.~Khan}, \bibinfo{author}{F.~Khan}, \bibinfo{author}{W.~Zuo}, \bibinfo{author}{R.~S.~M. Goh}, \bibinfo{author}{C.-M. Feng}, et~al.,
\newblock \bibinfo{title}{Sentence-level prompts benefit composed image retrieval},
\newblock in: \bibinfo{booktitle}{The Twelfth International Conference on Learning Representations}, \bibinfo{year}{2024}.
\bibitem[{Li et~al.(2022)Li, Li, Xiong, and Hoi}]{li2022blip}
\bibinfo{author}{J.~Li}, \bibinfo{author}{D.~Li}, \bibinfo{author}{C.~Xiong}, \bibinfo{author}{S.~Hoi},
\newblock \bibinfo{title}{Blip: Bootstrapping language-image pre-training for unified vision-language understanding and generation},
\newblock in: \bibinfo{booktitle}{International conference on machine learning}, \bibinfo{organization}{PMLR}, \bibinfo{year}{2022}, pp. \bibinfo{pages}{12888--12900}.
\bibitem[{Wen et~al.(2023)Wen, Zhang, Song, Wei, and Nie}]{wen2023target}
\bibinfo{author}{H.~Wen}, \bibinfo{author}{X.~Zhang}, \bibinfo{author}{X.~Song}, \bibinfo{author}{Y.~Wei}, \bibinfo{author}{L.~Nie},
\newblock \bibinfo{title}{Target-guided composed image retrieval},
\newblock in: \bibinfo{booktitle}{Proceedings of the 31st ACM International Conference on Multimedia}, \bibinfo{year}{2023}, pp. \bibinfo{pages}{915--923}.
\bibitem[{Delmas et~al.(2022)Delmas, de~Rezende, Csurka, and Larlus}]{delmas2022artemis}
\bibinfo{author}{G.~Delmas}, \bibinfo{author}{R.~S. de~Rezende}, \bibinfo{author}{G.~Csurka}, \bibinfo{author}{D.~Larlus},
\newblock \bibinfo{title}{Artemis: Attention-based retrieval with text-explicit matching and implicit similarity},
\newblock \bibinfo{journal}{ICLR}  (\bibinfo{year}{2022}).
\bibitem[{Jiang et~al.(2024)Jiang, Wang, Li, Wu, Hu, and Qian}]{jiang2024cala}
\bibinfo{author}{X.~Jiang}, \bibinfo{author}{Y.~Wang}, \bibinfo{author}{M.~Li}, \bibinfo{author}{Y.~Wu}, \bibinfo{author}{B.~Hu}, \bibinfo{author}{X.~Qian},
\newblock \bibinfo{title}{Cala: Complementary association learning for augmenting comoposed image retrieval},
\newblock in: \bibinfo{booktitle}{Proceedings of the 47th International ACM SIGIR Conference on Research and Development in Information Retrieval}, \bibinfo{year}{2024}, pp. \bibinfo{pages}{2177--2187}.
\bibitem[{Liu et~al.(2021)Liu, Rodriguez-Opazo, Teney, and Gould}]{liu2021cirr}
\bibinfo{author}{Z.~Liu}, \bibinfo{author}{C.~Rodriguez-Opazo}, \bibinfo{author}{D.~Teney}, \bibinfo{author}{S.~Gould},
\newblock \bibinfo{title}{Image retrieval on real-life images with pre-trained vision-and-language models},
\newblock in: \bibinfo{booktitle}{Proceedings of the IEEE/CVF International Conference on Computer Vision}, \bibinfo{year}{2021}, pp. \bibinfo{pages}{2125--2134}.
\bibitem[{Lin et~al.(2014)Lin, Maire, Belongie, Hays, Perona, Ramanan, Doll{\'a}r, and Zitnick}]{lin2014microsoft}
\bibinfo{author}{T.-Y. Lin}, \bibinfo{author}{M.~Maire}, \bibinfo{author}{S.~Belongie}, \bibinfo{author}{J.~Hays}, \bibinfo{author}{P.~Perona}, \bibinfo{author}{D.~Ramanan}, \bibinfo{author}{P.~Doll{\'a}r}, \bibinfo{author}{C.~L. Zitnick},
\newblock \bibinfo{title}{Microsoft coco: Common objects in context},
\newblock in: \bibinfo{booktitle}{Computer Vision--ECCV 2014: 13th European Conference, Zurich, Switzerland, September 6-12, 2014, Proceedings, Part V 13}, \bibinfo{organization}{Springer}, \bibinfo{year}{2014}, pp. \bibinfo{pages}{740--755}.
\bibitem[{Schuhmann et~al.(2022)Schuhmann, Beaumont, Vencu, Gordon, Wightman, Cherti, Coombes, Katta, Mullis, Wortsman et~al.}]{schuhmann2022laion5b}
\bibinfo{author}{C.~Schuhmann}, \bibinfo{author}{R.~Beaumont}, \bibinfo{author}{R.~Vencu}, \bibinfo{author}{C.~Gordon}, \bibinfo{author}{R.~Wightman}, \bibinfo{author}{M.~Cherti}, \bibinfo{author}{T.~Coombes}, \bibinfo{author}{A.~Katta}, \bibinfo{author}{C.~Mullis}, \bibinfo{author}{M.~Wortsman}, et~al.,
\newblock \bibinfo{title}{Laion-5b: An open large-scale dataset for training next generation image-text models},
\newblock \bibinfo{journal}{Advances in Neural Information Processing Systems} \bibinfo{volume}{35} (\bibinfo{year}{2022}) \bibinfo{pages}{25278--25294}.
\bibitem[{Gal et~al.(????)Gal, Alaluf, Atzmon, Patashnik, Bermano, Chechik, and Cohen-or}]{galimage}
\bibinfo{author}{R.~Gal}, \bibinfo{author}{Y.~Alaluf}, \bibinfo{author}{Y.~Atzmon}, \bibinfo{author}{O.~Patashnik}, \bibinfo{author}{A.~H. Bermano}, \bibinfo{author}{G.~Chechik}, \bibinfo{author}{D.~Cohen-or},
\newblock \bibinfo{title}{An image is worth one word: Personalizing text-to-image generation using textual inversion},
\newblock in: \bibinfo{booktitle}{The Eleventh International Conference on Learning Representations}, ????
\bibitem[{Gu et~al.(2023)Gu, Chun, Kim, Jun, Kang, and Yun}]{gu2023compodiff}
\bibinfo{author}{G.~Gu}, \bibinfo{author}{S.~Chun}, \bibinfo{author}{W.~Kim}, \bibinfo{author}{H.~Jun}, \bibinfo{author}{Y.~Kang}, \bibinfo{author}{S.~Yun},
\newblock \bibinfo{title}{Compodiff: Versatile composed image retrieval with latent diffusion},
\newblock \bibinfo{journal}{arXiv preprint arXiv:2303.11916}  (\bibinfo{year}{2023}).
\bibitem[{Levy et~al.(2024)Levy, Ben-Ari, Darshan, and Lischinski}]{levy2024data}
\bibinfo{author}{M.~Levy}, \bibinfo{author}{R.~Ben-Ari}, \bibinfo{author}{N.~Darshan}, \bibinfo{author}{D.~Lischinski},
\newblock \bibinfo{title}{Data roaming and quality assessment for composed image retrieval},
\newblock in: \bibinfo{booktitle}{Proceedings of the AAAI Conference on Artificial Intelligence}, volume~\bibinfo{volume}{38}, \bibinfo{year}{2024}, pp. \bibinfo{pages}{2991--2999}.
\bibitem[{Ventura et~al.(2024)Ventura, Yang, Schmid, and Varol}]{ventura2024covr}
\bibinfo{author}{L.~Ventura}, \bibinfo{author}{A.~Yang}, \bibinfo{author}{C.~Schmid}, \bibinfo{author}{G.~Varol},
\newblock \bibinfo{title}{Covr-2: Automatic data construction for composed video retrieval},
\newblock \bibinfo{journal}{IEEE Transactions on Pattern Analysis and Machine Intelligence}  (\bibinfo{year}{2024}).
\bibitem[{Rombach et~al.(2022)Rombach, Blattmann, Lorenz, Esser, and Ommer}]{rombach2022stablediffusion}
\bibinfo{author}{R.~Rombach}, \bibinfo{author}{A.~Blattmann}, \bibinfo{author}{D.~Lorenz}, \bibinfo{author}{P.~Esser}, \bibinfo{author}{B.~Ommer},
\newblock \bibinfo{title}{High-resolution image synthesis with latent diffusion models},
\newblock in: \bibinfo{booktitle}{Proceedings of the IEEE/CVF conference on computer vision and pattern recognition}, \bibinfo{year}{2022}, pp. \bibinfo{pages}{10684--10695}.
\bibitem[{Hertz et~al.(2022)Hertz, Mokady, Tenenbaum, Aberman, Pritch, and Cohen-Or}]{hertz2022prompttoprompt}
\bibinfo{author}{A.~Hertz}, \bibinfo{author}{R.~Mokady}, \bibinfo{author}{J.~Tenenbaum}, \bibinfo{author}{K.~Aberman}, \bibinfo{author}{Y.~Pritch}, \bibinfo{author}{D.~Cohen-Or}, \bibinfo{title}{Prompt-to-prompt image editing with cross attention control}, \bibinfo{year}{2022}. \URLprefix \url{https://arxiv.org/abs/2208.01626}. \href{http://arxiv.org/abs/2208.01626}{{\tt arXiv:2208.01626}}.
\bibitem[{Han et~al.(2022)Han, Yu, Zhu, Zhang, Song, and Xiang}]{han2022fashionvil}
\bibinfo{author}{X.~Han}, \bibinfo{author}{L.~Yu}, \bibinfo{author}{X.~Zhu}, \bibinfo{author}{L.~Zhang}, \bibinfo{author}{Y.-Z. Song}, \bibinfo{author}{T.~Xiang},
\newblock \bibinfo{title}{Fashionvil: Fashion-focused vision-and-language representation learning},
\newblock in: \bibinfo{booktitle}{European conference on computer vision}, \bibinfo{organization}{Springer}, \bibinfo{year}{2022}, pp. \bibinfo{pages}{634--651}.
\bibitem[{Han et~al.(2023)Han, Zhu, Yu, Zhang, Song, and Xiang}]{han2023fame}
\bibinfo{author}{X.~Han}, \bibinfo{author}{X.~Zhu}, \bibinfo{author}{L.~Yu}, \bibinfo{author}{L.~Zhang}, \bibinfo{author}{Y.-Z. Song}, \bibinfo{author}{T.~Xiang},
\newblock \bibinfo{title}{Fame-vil: Multi-tasking vision-language model for heterogeneous fashion tasks},
\newblock in: \bibinfo{booktitle}{Proceedings of the IEEE/CVF Conference on Computer Vision and Pattern Recognition}, \bibinfo{year}{2023}, pp. \bibinfo{pages}{2669--2680}.
\bibitem[{Brown et~al.(2020)Brown, Mann, Ryder, Subbiah, Kaplan, Dhariwal, Neelakantan, Shyam, Sastry, Askell et~al.}]{brown2020language}
\bibinfo{author}{T.~Brown}, \bibinfo{author}{B.~Mann}, \bibinfo{author}{N.~Ryder}, \bibinfo{author}{M.~Subbiah}, \bibinfo{author}{J.~D. Kaplan}, \bibinfo{author}{P.~Dhariwal}, \bibinfo{author}{A.~Neelakantan}, \bibinfo{author}{P.~Shyam}, \bibinfo{author}{G.~Sastry}, \bibinfo{author}{A.~Askell}, et~al.,
\newblock \bibinfo{title}{Language models are few-shot learners},
\newblock \bibinfo{journal}{Advances in neural information processing systems} \bibinfo{volume}{33} (\bibinfo{year}{2020}) \bibinfo{pages}{1877--1901}.
\bibitem[{Dubey et~al.(2024)Dubey, Jauhri, Pandey, Kadian, Al-Dahle, Letman, Mathur, Schelten, Yang, Fan et~al.}]{dubey2024llama}
\bibinfo{author}{A.~Dubey}, \bibinfo{author}{A.~Jauhri}, \bibinfo{author}{A.~Pandey}, \bibinfo{author}{A.~Kadian}, \bibinfo{author}{A.~Al-Dahle}, \bibinfo{author}{A.~Letman}, \bibinfo{author}{A.~Mathur}, \bibinfo{author}{A.~Schelten}, \bibinfo{author}{A.~Yang}, \bibinfo{author}{A.~Fan}, et~al.,
\newblock \bibinfo{title}{The llama 3 herd of models},
\newblock \bibinfo{journal}{arXiv preprint arXiv:2407.21783}  (\bibinfo{year}{2024}).
\bibitem[{Liu et~al.(2024)Liu, Li, Wu, and Lee}]{liu2024visual}
\bibinfo{author}{H.~Liu}, \bibinfo{author}{C.~Li}, \bibinfo{author}{Q.~Wu}, \bibinfo{author}{Y.~J. Lee},
\newblock \bibinfo{title}{Visual instruction tuning},
\newblock \bibinfo{journal}{Advances in neural information processing systems} \bibinfo{volume}{36} (\bibinfo{year}{2024}).
\bibitem[{Achiam et~al.(2023)Achiam, Adler, Agarwal, Ahmad, Akkaya, Aleman, Almeida, Altenschmidt, Altman, Anadkat et~al.}]{achiam2023gpt}
\bibinfo{author}{J.~Achiam}, \bibinfo{author}{S.~Adler}, \bibinfo{author}{S.~Agarwal}, \bibinfo{author}{L.~Ahmad}, \bibinfo{author}{I.~Akkaya}, \bibinfo{author}{F.~L. Aleman}, \bibinfo{author}{D.~Almeida}, \bibinfo{author}{J.~Altenschmidt}, \bibinfo{author}{S.~Altman}, \bibinfo{author}{S.~Anadkat}, et~al.,
\newblock \bibinfo{title}{Gpt-4 technical report},
\newblock \bibinfo{journal}{arXiv preprint arXiv:2303.08774}  (\bibinfo{year}{2023}).
\bibitem[{Chen et~al.(2024)Chen, Wu, Wang, Su, Chen, Xing, Zhong, Zhang, Zhu, Lu et~al.}]{chen2024internvl}
\bibinfo{author}{Z.~Chen}, \bibinfo{author}{J.~Wu}, \bibinfo{author}{W.~Wang}, \bibinfo{author}{W.~Su}, \bibinfo{author}{G.~Chen}, \bibinfo{author}{S.~Xing}, \bibinfo{author}{M.~Zhong}, \bibinfo{author}{Q.~Zhang}, \bibinfo{author}{X.~Zhu}, \bibinfo{author}{L.~Lu}, et~al.,
\newblock \bibinfo{title}{Internvl: Scaling up vision foundation models and aligning for generic visual-linguistic tasks},
\newblock in: \bibinfo{booktitle}{Proceedings of the IEEE/CVF Conference on Computer Vision and Pattern Recognition}, \bibinfo{year}{2024}, pp. \bibinfo{pages}{24185--24198}.
\bibitem[{Wang et~al.(2022)Wang, Yang, Hu, Li, Lin, Gan, Liu, Liu, and Wang}]{wang2022git}
\bibinfo{author}{J.~Wang}, \bibinfo{author}{Z.~Yang}, \bibinfo{author}{X.~Hu}, \bibinfo{author}{L.~Li}, \bibinfo{author}{K.~Lin}, \bibinfo{author}{Z.~Gan}, \bibinfo{author}{Z.~Liu}, \bibinfo{author}{C.~Liu}, \bibinfo{author}{L.~Wang},
\newblock \bibinfo{title}{Git: A generative image-to-text transformer for vision and language},
\newblock \bibinfo{journal}{arXiv preprint arXiv:2205.14100}  (\bibinfo{year}{2022}).
\bibitem[{Gan et~al.(2022)Gan, Li, Li, Wang, Liu, Gao et~al.}]{gan2022vision}
\bibinfo{author}{Z.~Gan}, \bibinfo{author}{L.~Li}, \bibinfo{author}{C.~Li}, \bibinfo{author}{L.~Wang}, \bibinfo{author}{Z.~Liu}, \bibinfo{author}{J.~Gao}, et~al.,
\newblock \bibinfo{title}{Vision-language pre-training: Basics, recent advances, and future trends},
\newblock \bibinfo{journal}{Foundations and Trends{\textregistered} in Computer Graphics and Vision} \bibinfo{volume}{14} (\bibinfo{year}{2022}) \bibinfo{pages}{163--352}.
\bibitem[{Antol et~al.(2015)Antol, Agrawal, Lu, Mitchell, Batra, Zitnick, and Parikh}]{antol2015vqa}
\bibinfo{author}{S.~Antol}, \bibinfo{author}{A.~Agrawal}, \bibinfo{author}{J.~Lu}, \bibinfo{author}{M.~Mitchell}, \bibinfo{author}{D.~Batra}, \bibinfo{author}{C.~L. Zitnick}, \bibinfo{author}{D.~Parikh},
\newblock \bibinfo{title}{Vqa: Visual question answering},
\newblock in: \bibinfo{booktitle}{Proceedings of the IEEE international conference on computer vision}, \bibinfo{year}{2015}, pp. \bibinfo{pages}{2425--2433}.
\bibitem[{Yang et~al.(2020)Yang, Zhang, Jin, Liu, Wu, Tan, Xie, Wang, and Wang}]{Xuewen2020FACAD}
\bibinfo{author}{X.~Yang}, \bibinfo{author}{H.~Zhang}, \bibinfo{author}{D.~Jin}, \bibinfo{author}{Y.~Liu}, \bibinfo{author}{C.-H. Wu}, \bibinfo{author}{J.~Tan}, \bibinfo{author}{D.~Xie}, \bibinfo{author}{J.~Wang}, \bibinfo{author}{X.~Wang},
\newblock \bibinfo{title}{Fashion captioning: Towards generating accurate descriptions with semantic rewards},
\newblock in: \bibinfo{booktitle}{ECCV}, \bibinfo{year}{2020}.
\bibitem[{Liu et~al.(2024)Liu, Li, Li, Li, Zhang, Shen, and Lee}]{liu2024llavanext}
\bibinfo{author}{H.~Liu}, \bibinfo{author}{C.~Li}, \bibinfo{author}{Y.~Li}, \bibinfo{author}{B.~Li}, \bibinfo{author}{Y.~Zhang}, \bibinfo{author}{S.~Shen}, \bibinfo{author}{Y.~J. Lee}, \bibinfo{title}{Llava-next: Improved reasoning, ocr, and world knowledge}, \bibinfo{year}{2024}. \URLprefix \url{https://llava-vl.github.io/blog/2024-01-30-llava-next/}.
\bibitem[{Li et~al.(2024{\natexlab{a}})Li, Zhang, Guo, Zhang, Li, Zhang, Zhang, Zhang, Li, Liu et~al.}]{li2024llavaonevision}
\bibinfo{author}{B.~Li}, \bibinfo{author}{Y.~Zhang}, \bibinfo{author}{D.~Guo}, \bibinfo{author}{R.~Zhang}, \bibinfo{author}{F.~Li}, \bibinfo{author}{H.~Zhang}, \bibinfo{author}{K.~Zhang}, \bibinfo{author}{P.~Zhang}, \bibinfo{author}{Y.~Li}, \bibinfo{author}{Z.~Liu}, et~al.,
\newblock \bibinfo{title}{Llava-onevision: Easy visual task transfer},
\newblock \bibinfo{journal}{arXiv preprint arXiv:2408.03326}  (\bibinfo{year}{2024}{\natexlab{a}}).
\bibitem[{Li et~al.(2024{\natexlab{b}})Li, Zhang, Zhang, Zhang, Li, Li, Ma, and Li}]{li2024llavanextinterleave}
\bibinfo{author}{F.~Li}, \bibinfo{author}{R.~Zhang}, \bibinfo{author}{H.~Zhang}, \bibinfo{author}{Y.~Zhang}, \bibinfo{author}{B.~Li}, \bibinfo{author}{W.~Li}, \bibinfo{author}{Z.~Ma}, \bibinfo{author}{C.~Li},
\newblock \bibinfo{title}{Llava-next-interleave: Tackling multi-image, video, and 3d in large multimodal models},
\newblock \bibinfo{journal}{arXiv preprint arXiv:2407.07895}  (\bibinfo{year}{2024}{\natexlab{b}}).
\bibitem[{Shen et~al.(2024)Shen, Xiong, Zhao, Wu, Chen, Zhu, Liu, Xiao, Varadarajan, Bordes et~al.}]{shen2024longvu}
\bibinfo{author}{X.~Shen}, \bibinfo{author}{Y.~Xiong}, \bibinfo{author}{C.~Zhao}, \bibinfo{author}{L.~Wu}, \bibinfo{author}{J.~Chen}, \bibinfo{author}{C.~Zhu}, \bibinfo{author}{Z.~Liu}, \bibinfo{author}{F.~Xiao}, \bibinfo{author}{B.~Varadarajan}, \bibinfo{author}{F.~Bordes}, et~al.,
\newblock \bibinfo{title}{Longvu: Spatiotemporal adaptive compression for long video-language understanding},
\newblock \bibinfo{journal}{arXiv preprint arXiv:2410.17434}  (\bibinfo{year}{2024}).
\bibitem[{Li et~al.(2024)Li, Yuan, Liu, Tang, Wang, Qin, Zhu, and Zhang}]{li2024tokenpacker}
\bibinfo{author}{W.~Li}, \bibinfo{author}{Y.~Yuan}, \bibinfo{author}{J.~Liu}, \bibinfo{author}{D.~Tang}, \bibinfo{author}{S.~Wang}, \bibinfo{author}{J.~Qin}, \bibinfo{author}{J.~Zhu}, \bibinfo{author}{L.~Zhang},
\newblock \bibinfo{title}{Tokenpacker: Efficient visual projector for multimodal llm},
\newblock \bibinfo{journal}{arXiv preprint arXiv:2407.02392}  (\bibinfo{year}{2024}).
\bibitem[{Han et~al.(2017)Han, Wu, Huang, Zhang, Zhu, Li, Zhao, and Davis}]{han2017automatic}
\bibinfo{author}{X.~Han}, \bibinfo{author}{Z.~Wu}, \bibinfo{author}{P.~X. Huang}, \bibinfo{author}{X.~Zhang}, \bibinfo{author}{M.~Zhu}, \bibinfo{author}{Y.~Li}, \bibinfo{author}{Y.~Zhao}, \bibinfo{author}{L.~S. Davis},
\newblock \bibinfo{title}{Automatic spatially-aware fashion concept discovery},
\newblock in: \bibinfo{booktitle}{Proceedings of the IEEE international conference on computer vision}, \bibinfo{year}{2017}, pp. \bibinfo{pages}{1463--1471}.
\bibitem[{Jiang et~al.(2022)Jiang, Yang, Qiu, Wu, Loy, and Liu}]{jiang2022text2human}
\bibinfo{author}{Y.~Jiang}, \bibinfo{author}{S.~Yang}, \bibinfo{author}{H.~Qiu}, \bibinfo{author}{W.~Wu}, \bibinfo{author}{C.~C. Loy}, \bibinfo{author}{Z.~Liu},
\newblock \bibinfo{title}{Text2human: Text-driven controllable human image generation},
\newblock \bibinfo{journal}{ACM Transactions on Graphics (TOG)} \bibinfo{volume}{41} (\bibinfo{year}{2022}) \bibinfo{pages}{1--11}. \DOIprefix\doi{10.1145/3528223.3530104}.
\bibitem[{Liu et~al.(2016)Liu, Luo, Qiu, Wang, and Tang}]{liu2016deepfashion}
\bibinfo{author}{Z.~Liu}, \bibinfo{author}{P.~Luo}, \bibinfo{author}{S.~Qiu}, \bibinfo{author}{X.~Wang}, \bibinfo{author}{X.~Tang},
\newblock \bibinfo{title}{Deepfashion: Powering robust clothes recognition and retrieval with rich annotations},
\newblock in: \bibinfo{booktitle}{Proceedings of the IEEE conference on computer vision and pattern recognition}, \bibinfo{year}{2016}, pp. \bibinfo{pages}{1096--1104}.
\bibitem[{Sung et~al.(2022)Sung, Cho, and Bansal}]{sung2022vl}
\bibinfo{author}{Y.-L. Sung}, \bibinfo{author}{J.~Cho}, \bibinfo{author}{M.~Bansal},
\newblock \bibinfo{title}{Vl-adapter: Parameter-efficient transfer learning for vision-and-language tasks},
\newblock in: \bibinfo{booktitle}{Proceedings of the IEEE/CVF conference on computer vision and pattern recognition}, \bibinfo{year}{2022}, pp. \bibinfo{pages}{5227--5237}.
\bibitem[{Waswani et~al.(2017)Waswani, Shazeer, Parmar, Uszkoreit, Jones, Gomez, Kaiser, and Polosukhin}]{waswani2017attention}
\bibinfo{author}{A.~Waswani}, \bibinfo{author}{N.~Shazeer}, \bibinfo{author}{N.~Parmar}, \bibinfo{author}{J.~Uszkoreit}, \bibinfo{author}{L.~Jones}, \bibinfo{author}{A.~Gomez}, \bibinfo{author}{L.~Kaiser}, \bibinfo{author}{I.~Polosukhin},
\newblock \bibinfo{title}{Attention is all you need},
\newblock in: \bibinfo{booktitle}{NIPS}, \bibinfo{year}{2017}.
\bibitem[{Hendrycks and Gimpel(2016)}]{hendrycks2016gaussian}
\bibinfo{author}{D.~Hendrycks}, \bibinfo{author}{K.~Gimpel},
\newblock \bibinfo{title}{Gaussian error linear units (gelus)},
\newblock \bibinfo{journal}{arXiv preprint arXiv:1606.08415}  (\bibinfo{year}{2016}).
\bibitem[{Tolstikhin et~al.(2021)Tolstikhin, Houlsby, Kolesnikov, Beyer, Zhai, Unterthiner, Yung, Steiner, Keysers, Uszkoreit et~al.}]{tolstikhin2021mlp}
\bibinfo{author}{I.~O. Tolstikhin}, \bibinfo{author}{N.~Houlsby}, \bibinfo{author}{A.~Kolesnikov}, \bibinfo{author}{L.~Beyer}, \bibinfo{author}{X.~Zhai}, \bibinfo{author}{T.~Unterthiner}, \bibinfo{author}{J.~Yung}, \bibinfo{author}{A.~Steiner}, \bibinfo{author}{D.~Keysers}, \bibinfo{author}{J.~Uszkoreit}, et~al.,
\newblock \bibinfo{title}{Mlp-mixer: An all-mlp architecture for vision},
\newblock \bibinfo{journal}{Advances in neural information processing systems} \bibinfo{volume}{34} (\bibinfo{year}{2021}) \bibinfo{pages}{24261--24272}.

\end{thebibliography}




\end{document}